\documentclass[sigconf,nonacm]{acmart}

\setcopyright{none}

\AtBeginDocument{%
}

\usepackage{subcaption}

\usepackage{algorithm}
\usepackage{algorithmic}
\usepackage{tikz}

\begin{document}

%% ---------- Title ----------
\title{CoMIC: Collaborative Memory and Insights Circulation for Long-Horizon LLM Agents in Web Cloud-Edge Systems}

%% ---------- Authors (hidden in anonymous mode, but fill them in) ----------
% \author{First Author}
% \email{first.author@example.com}
% \affiliation{%
%   \institution{Institution}
%   \city{City}
%   \country{Country}}

% \author{Second Author}
% \email{second.author@example.com}
% \affiliation{%
%   \institution{Institution}
%   \city{City}
%   \country{Country}}

% \renewcommand{\shortauthors}{Author et al.}

\author{Yannan Wang}
\authornote{Both authors contributed equally to this research.}
\authornote{Visiting PhD student at WMG, University of Warwick.}
\email{yannanwang@bjtu.edu.cn}
\affiliation{%
  \department{School of Computer Science and Technology}
  \institution{Beijing Jiaotong University}
  \city{Beijing}
  \postcode{100044}
  \country{China}}

\author{Longli Yang}
\authornotemark[1]
\email{yanglongli26@mails.ucas.ac.cn}
\affiliation{%
  \department{Hangzhou Institute for Advanced Study}
  \institution{University of the Chinese Academy of Sciences}
  \city{Hangzhou}
  \postcode{310024}
  \country{China}}

\author{Zhen Liu}
\authornote{Corresponding author.}
\email{zhliu@bjtu.edu.cn}
\affiliation{%
  \institution{Beijing Jiaotong University}
  \city{Beijing}
  \postcode{100044}
  \country{China}}

\author{Abhishek Kumar}
\email{Abhishek.Kumar.3@warwick.ac.uk}
\affiliation{%
  \department{WMG}
  \institution{University of Warwick}
  \city{Coventry}
  \postcode{CV4 7AL}
  \country{United Kingdom}}

\author{Carsten Maple}
\email{CM@warwick.ac.uk}
\affiliation{%
  \department{WMG}
  \institution{University of Warwick}
  \city{Coventry}
  \postcode{CV4 7AL}
  \country{United Kingdom}}
\additionalaffiliation{%
  \institution{The Alan Turing Institute}
  \city{London}
  \country{United Kingdom}}

\renewcommand{\shortauthors}{Wang et al.}

%% ---------- Abstract ----------
\begin{abstract}
LLM-based Web agents are increasingly deployed on edge servers to carry out long-horizon tasks on behalf of users with low latency. However, the lightweight models that fit on edge servers often struggle with long-horizon tasks that require persistent memory, subgoal tracking, and reflection. Fine-tuning edge models after deployment is costly and difficult to scale across heterogeneous nodes, while purely local execution confines each agent to its own isolated experience. We propose \textsc{CoMIC}, a parameter-update-free cloud-edge framework for Collaborative Memory and Insights Circulation. Following the paradigm of ``Act at the Edge, Reflect in the Cloud'', edge agents execute locally with subgoal-oriented hierarchical memory and selective re-expansion of relevant histories, while a cloud-side LLM critic asynchronously evaluates the trajectories uploaded through a Web gateway, admits reusable experience, and circulates cross-agent guidance indexed by semantic subgoal identifiers back to all edge agents. On five long-horizon benchmarks, \textsc{CoMIC} enables weak edge agents to achieve non-zero success on tasks where a standard memory-based agent fails entirely, and improves their task progress over both the standard agent and purely local execution while reducing context consumption relative to the standard agent, all without updating model parameters. When stronger heterogeneous agents join the edge, the weak agents further improve in success, progress, and grounding with lower context consumption.
\end{abstract}

%% ---------- CCS concepts (generate at https://dl.acm.org/ccs and replace) ----------

% \ccsdesc[500]{Information systems~World Wide Web}

%% ---------- Keywords ----------

\maketitle

%% ============================================================
%% Main text (8 pages max)
%% ============================================================
\section{Introduction}
The Web is shifting from a space where users search and browse by themselves to one where AI assistants accomplish tasks on their behalf~\cite{Shao2025}. Benefiting from recent breakthroughs in large language models (LLMs), LLM-based agents have shown great potential in autonomous decision-making~\cite{Borzilov2025, Zhu2025}, enabling Web applications to translate rapid advances in AI into more personalized and convenient user experiences~\cite{Wang2024, Xi2025Rise}. Such agents have proven effective in helping users handle massive Web content and tasks, including long-horizon tasks such as Web browsing~\cite{Lai2024}, content search~\cite{Sharma2024}, online shopping~\cite{Nie2024}, and travel planning~\cite{Xie2024}. These tasks involve dozens of interdependent decisions, require persistent tracking of intermediate progress, and call for reflection on failures~\cite{Yao2022, Hu2025Hiagent}.

\begin{figure}[htbp]
  \centering
  \includegraphics[width=\linewidth]{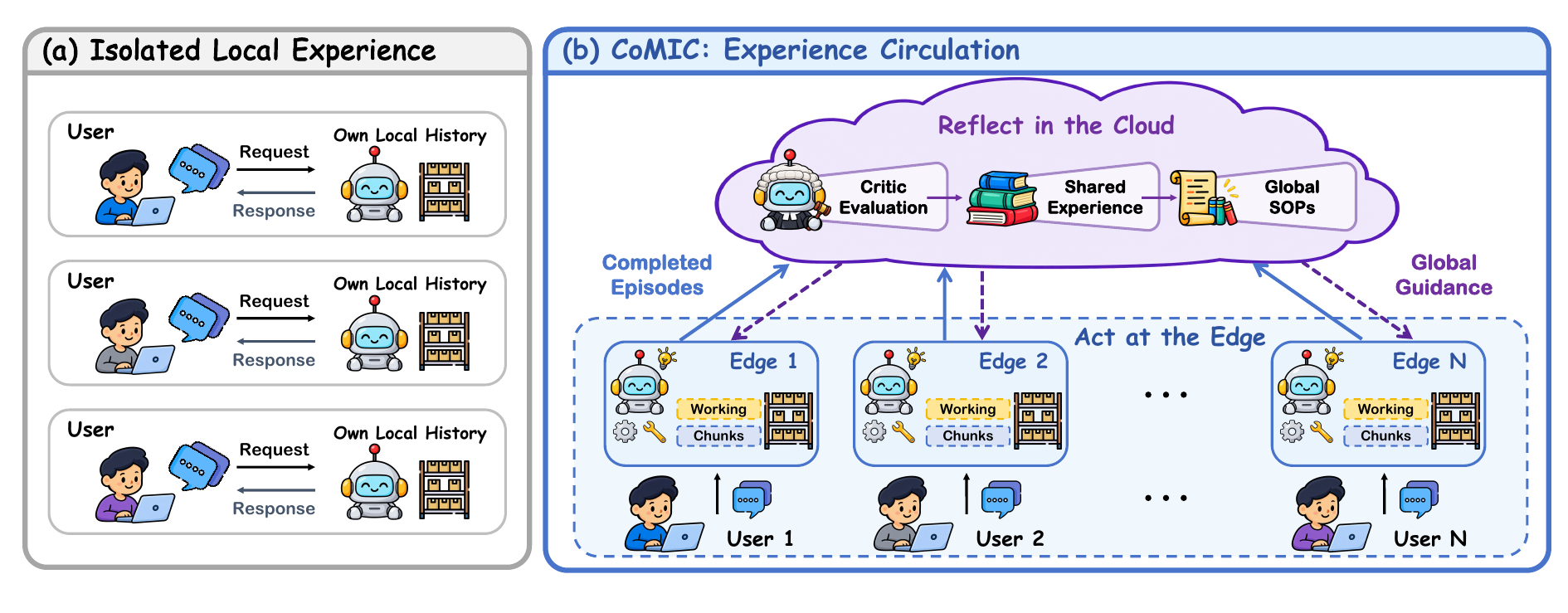}
  \caption{(a) Isolated local experience of Web edge agents; (b) experience circulation in \textsc{CoMIC}.}
  \label{fig:cloud-edge-end-scenario}
\end{figure}

To meet users' demand for low latency when performing Web tasks, Web infrastructure is pushing these agents toward the network edge. Deploying lightweight LLM agents on edge servers to serve users' end devices has become a practical solution for bringing autonomous capabilities closer to users~\cite{Qu2025}. However, as illustrated in Figure~\ref{fig:cloud-edge-end-scenario}(a), constrained by the limited computational resources of edge servers and the limited context capacity of lightweight models, edge-deployed LLM agents exhibit restricted reasoning capability on complex long-horizon tasks~\cite{Yao2025}. Although continuously fine-tuning models on task experience is a conventional approach, performing frequent parameter updates across highly diverse tasks and heterogeneous edge nodes incurs substantial computational overhead, making it difficult to scale in practice~\cite{Zeng2024}. Therefore, optimizing text-based memory interaction mechanisms without parameter updates offers a more scalable alternative~\cite{Zhang2025}. However, classic memory paradigms that concatenate entire interaction histories into prompts suffer from context explosion and high token consumption~\cite{Yao2022, Hu2025Hiagent}. Recent studies mitigate this by decomposing memory into cross-trial and in-trial (working) memories~\cite{Xi2025, Zhang2025Learn, Li2025}, yet maintaining and retrieving cross-trial memories accumulated over many tasks introduces considerable computational and storage overheads, making them better suited to resource-rich environments than to edge servers. Moreover, purely local execution confines resource-bounded edge agents to isolated experiences, lacking the cross-agent sharing required to support complex tasks efficiently.

The constraints faced by edge-deployed agents resemble the cognitive limitations of human individuals: with limited memory and cognitive resources, no single individual can master all relevant knowledge alone. To cope with this, human groups have developed a fundamental strategy~\cite{Wegner1987}: they accomplish complex tasks collaboratively through distributed memory and shared knowledge. Inspired by this observation, we propose \textbf{Co}llaborative \textbf{M}emory and \textbf{I}nsights \textbf{C}irculation (\textsc{CoMIC}), a parameter-update-free cloud-edge collaborative framework for memory-enhanced edge LLM agents. Operating under an ``Act at the Edge, Reflect in the Cloud'' paradigm(Figure~\ref{fig:cloud-edge-end-scenario}(b)), \textsc{CoMIC} offloads computationally intensive tasks, such as cross-trial long-term memory maintenance and complex logical reflection, to the cloud. Meanwhile, edge nodes maintain only lightweight, selectively expandable memories for immediate decision-making, thereby improving the capability of agents deployed at the Web edge to handle complex long-horizon tasks under resource constraints. In summary, our contributions are as follows.
\begin{itemize}
  \item We design \textsc{CoMIC}, a cloud-edge collaborative framework for lightweight LLM agents deployed on Web edge servers. Edge agents make subgoal-driven decisions autonomously by combining local hierarchical memory with global guidance from the cloud, and upload trajectories asynchronously over Web protocols to keep online execution unblocked, thereby establishing a cross-agent experience-circulation layer within Web infrastructure.
  \item A cloud-side LLM serves as a global critic that independently evaluates individual trajectories asynchronously uploaded by edge agents and distills reusable experiences through admission control. Using semantic subgoal identifiers as indexing keys, it then aggregates experiences of compatible subgoals from different edge agents into global guidance and dispatches it to edge agents facing matching subgoals, all without updating model parameters.
  \item Experiments on five long-horizon benchmarks show that, when only weak LLM agents are deployed at the Web edge, \textsc{CoMIC} enables them to achieve non-zero success on tasks where a standard memory-based agent fails entirely, and improves the overall success rate and subgoal-level progress rate compared with purely local execution. Although weak agents produce few high-quality trajectories, the cloud critic can still extract subgoal-level experience that drives task progress, and introducing stronger heterogeneous edge agents supplies more high-quality trajectories, yielding further gains with lower context consumption.
\end{itemize}

\section{Related Work}
\subsection{LLM Agents on the Web}
LLM-based agents have become a key paradigm for Web services, performing multi-step reasoning, tool use, and interaction with Web environments on behalf of users~\cite{Wang2024, Xi2025Rise, Li2026DeepAgent}. To deliver such services closer to users, recent work runs LLMs and agents on resource-constrained platforms. WeInfer accelerates LLM inference inside Web browsers via WebGPU~\cite{Chen2025}, LaTune adaptively tunes runtime configurations for LLM inference on edge devices~\cite{Zhong2026}, and personal and on-device agents have been developed to assist users on their own devices~\cite{Li2024Personal, Chen2025Octo}. However, deploying capable LLMs on such platforms remains challenging, and the lightweight models that fit them fall short in complex reasoning~\cite{Shao2025, Li2024Personal}. Local resources alone thus cannot support complex long-horizon tasks.

\subsection{LLM Agents in Cloud--Edge Systems}
Mobile edge intelligence places computing resources at the network edge close to users, reducing service latency and keeping data local~\cite{Lin2021, Qu2025}. Building on this infrastructure, edge-cloud generative AI services have been extensively explored~\cite{Xu2024}. Industry has combined on-device foundation models with server-side models~\cite{April2025}, while compact models such as Gemini Nano and MobileLLM have been optimized for on-device use~\cite{Team2023, Liu2024}. Recent studies further improve LLM serving at the edge through inter-task KV-cache reuse~\cite{Luo2025} and cloud-edge request scheduling based on multi-agent reinforcement learning~\cite{Yao2025}. For LLM agents, DoT decomposes user queries into sub-tasks and allocates them between a local small model and a cloud LLM~\cite{Shao2025}. AdaSwitch enables a local agent to seek assistance from a cloud agent when it detects potential errors, and trains the local agent on the corrected trajectories~\cite{Sun2024}. EcoAgent and AdecPilot let the cloud generate plans or milestones that are then executed by edge agents~\cite{Yi2025, Li2026}. In addition, IntellectReq determines when edge models should request updated parameters from the cloud~\cite{Lv2024}. Consequently, the capability of edge agents in these systems either depends on the cloud's online processing of edge requests or requires updating the parameters of edge models.

\subsection{Memory and Experience Sharing for LLM Agents}
Memory enables LLM-based agents to accumulate knowledge and retrieve relevant past experience for decision-making~\cite{Zhang2025, Wang2024}. Within a task, agents compress or fold lengthy interaction histories into structured memories~\cite{Li2026DeepAgent, Lindenbauer2025}. Across tasks, they reflect on failures~\cite{Shinn2023}, distill insights~\cite{Zhao2024Expel}, or induce reusable workflows~\cite{Wang2025} from past trajectories. Recent studies further share experience among agents through cross-task experiential learning in multi-agent collaboration~\cite{Li2025} and hierarchical insight and interaction memories for multi-agent systems~\cite{Zhang2026}, and have shown that the quality of stored experience shapes subsequent agent behavior~\cite{Xiong2026}.

However, for edge-deployed LLM agents, limited computational resources hinder efficient reasoning and reflection, while constrained storage restricts the retention of long-term, multi-trial memories. As a result, edge LLM agents cannot accumulate and learn from experience as typical LLM agents do, which calls for cloud-edge collaboration that supports memory and experience sharing.

\section{Preliminaries}
\subsection{Cloud-Edge Deployment on the Web}
We consider a cloud-edge-end system on the Web. Task requests issued from end-user devices are served by a set of edge agents $\mathcal{E}=\{e^1,\dots,e^N\}$ deployed on Web edge servers. Each edge agent $e^i$ is driven by a lightweight LLM $\pi_{\theta^i}$ with parameters $\theta^i$ fixed after deployment, and backbone models may differ across agents. A cloud node hosts a more capable LLM and communicates with edge agents asynchronously over Web protocols (HTTP/WebSocket).

\subsection{Subgoal Episodes and Trajectories}\label{sec:prelim-traj}
Given a global task $g$, edge agent $e^i$ organizes its execution as a sequence of subgoal episodes, each corresponding to one subgoal, rather than a flat interaction history. In episode $t$, the agent pursues subgoal $g^i_t$, receives observation $o^i_t$, executes action $a^i_t$, and obtains outcome $r^i_t$. The trajectory of episode $t$ is summarized as
\begin{equation}
\hat{\tau}^i_t=(g,\ g^i_t,\ o^i_t,\ a^i_t,\ r^i_t),
\end{equation}
and the global trajectory is the ordered set of all subgoal episodes, $\tau^i_g=\{\hat{\tau}^i_1,\dots,\hat{\tau}^i_{T^i}\}$, where $T^i$ denotes the number of subgoals into which $e^i$ divides task $g$.

\subsection{Two Levels of Agent Memory}
We distinguish two levels of agent memory~\cite{Zhang2025}. In-trial (working) memory $H^i_t$ maintains the information required for the current task: the active subgoal episode retains its detailed action-observation records, while earlier completed episodes are compressed into summaries. Cross-trial memory accumulates experience that can be reused across tasks and agents, such as summaries, insights, and suggestions distilled from completed trajectories.

\section{Method}
In this section, we formally present \textsc{CoMIC}, a memory-centric cloud-edge collaborative framework. This framework improves the long-horizon decision-making capabilities of lightweight edge LLM agents by summarizing cross-agent experiences in the cloud and returning selected guidance to the edge agents.

\subsection{Framework Overview}\label{sec:overview}
\begin{figure*}[t]
  \centering
  \includegraphics[width=\linewidth]{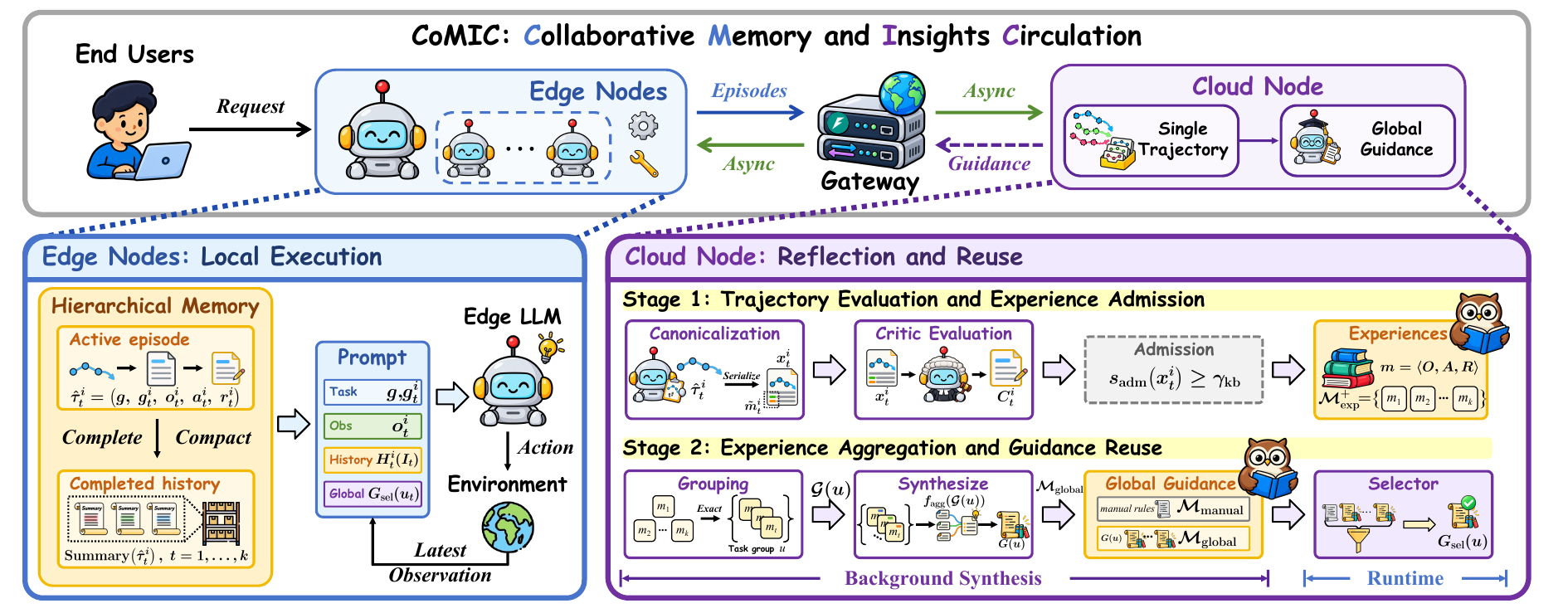}
  \caption{Overview of the \textsc{CoMIC} framework, including local execution at the edge and reflection and reuse in the cloud.}
  \label{fig:framework}
\end{figure*}

Figure~\ref{fig:framework} illustrates the workflow of \textsc{CoMIC}, which operates under an ``Act at the Edge, Reflect in the Cloud'' paradigm. Lightweight agents deployed on Web edge servers handle local decision-making and environment interaction, while the cloud performs experience reflection and cross-edge knowledge aggregation outside the online execution loop.

The workflow initiates at the end layer with a task request issued by a user. At the edge, the local agent organizes this request as a sequence of subgoal episodes, interacts with the environment, and records the resulting traces as structured trajectories. To ensure real-time responsiveness, trajectories are uploaded asynchronously without blocking local execution.

Upon receiving the uploaded trajectories, the cloud improves future decisions through two feedback mechanisms. The first is trajectory evaluation, in which an LLM acting as a global critic analyzes each completed subgoal episode and produces trajectory-grounded reflections that support evidence admission and later reuse. The second operates at a global level: instead of focusing on a single trajectory, the cloud distills reusable knowledge from admitted experiences accumulated across compatible agents and tasks, and returns it to the edge as \emph{Global Guidance}, the single advisory channel for subsequent decisions.

Edge agents and the cloud communicate over Web protocols through a gateway built on FastAPI and WebSocket. Completed subgoal episodes are uploaded together with their task and subgoal metadata, and the gateway dispatches each record according to its canonical stage and key. Aggregated guidance is broadcast back to the edge, where it is assembled into later decision prompts, and edge agents acknowledge the guidance they receive. Because both uploading and guidance synthesis proceed asynchronously, the online execution of edge agents is never blocked by cloud-side processing. Detailed designs of the edge memory framework, the cloud critic, and the collaboration mechanism are presented in \S\ref{sec:edge}, \S\ref{sec:cloud}, and \S\ref{sec:collab}, respectively.

\subsection{Edge Framework}\label{sec:edge}
The edge agent of \textsc{CoMIC} adopts a planning--execution mode for local tasks based on subgoal-oriented working memory, keeping immediate interactive steps on the edge while offloading global information processing to the cloud.

\subsubsection{Subgoal-Driven Execution}
For a global task $g$, the edge agent $e^i$ decomposes it into subgoal-driven episodes and records the global trajectory $\tau^i_g$ defined in \S\ref{sec:prelim-traj}. By strictly aligning each local decision with the active subgoal, the agent structurally reduces the reasoning burden on the edge.

\subsubsection{Hierarchical Memory and Summarizer}
The edge agent's local memory employs a two-level hierarchical structure. The \emph{working} memory holds the active subgoal episode and retains its detailed action--observation pairs to inform precise local decisions. Earlier completed episodes are stored in the history as memory \emph{chunks}, each compressed into an abstract summary when suitable; in compact symbolic PDDL domains such as Blocksworld and Gripper, concise predicate-level states are preserved to avoid losing object grounding. To retrieve detailed trajectories without overflowing the context limit, the system reconstructs historical contexts via selective re-expansion. Formally, given an index set $\mathcal{I}$ of previously summarized subgoals relevant to the current decision, the agent reconstructs the historical context $H^i_t$:
\begin{equation}\label{eq:history}
H^i_t(\mathcal{I})=\big\{\mathrm{Summary}(\hat{\tau}^i_t)\,\big|\,t\notin\mathcal{I}\big\}\cup\big\{\hat{\tau}^i_t\,\big|\,t\in\mathcal{I}\big\}.
\end{equation}
Reusable cloud guidance is kept outside both levels as a separate prompt-level advisory signal.

\begin{figure*}[t]
\centering
\captionsetup[subfigure]{labelformat=simple}
\begin{subfigure}[t]{0.48\textwidth}
\centering
\begin{minipage}[t]{\linewidth}
\color{gray}\scriptsize\ttfamily\raggedright
You are a cloud-side trajectory evaluation assistant.\\
Your job is to evaluate one execution trajectory and return exactly one JSON object.\\[0.5em]
Rules:\\
1. Use only the trajectory and optional context.\\
2. Do not invent facts or unstated outcomes.\\
3. If evidence is incomplete, lower self\_reported\_confidence.\\
4. Keep summary, insights, and suggestions trajectory-grounded.\\
5. Output JSON only.\\[0.5em]
Return keys:\\
\{"summary", "insights", "suggestions", "self\_reported\_confidence"\}\\[0.5em]
Task / Subgoals / Metadata: \{trajectory\_text\}\\
Optional Context: \{optional\_context\}
\end{minipage}
\caption{Cloud critic prompt.}
\label{prompt:cloud_critic}
\end{subfigure}
\hfill
\begin{subfigure}[t]{0.48\textwidth}
\centering
\begin{minipage}[t]{\linewidth}
\color{gray}\scriptsize\ttfamily\raggedright
Note: A subgoal is a milestone goal toward the final goal.\\
If an unfinished subgoal exists, output "Action: \{action\}".\\
If the previous subgoal has been completed, output\\
"Subgoal: \{subgoal\}\textbackslash nAction: \{action\}".\\[0.5em]
Instructions:\\
1. Do not output two consecutive subgoals.\\
2. Subgoal must be one line.\\
3. If an action fails, use "check valid actions".\\
4. Use "retrieve(subgoal\_id)" only when hidden detailed history is needed.\\[0.5em]
\{examples\}\\
Goal: \{goal\}\\
Global Guidance: \{global SOP / insight / suggestion\}\\
\{serialized\_history\}\\
Action:
\end{minipage}
\caption{Edge-agent prompt.}
\label{prompt:edge_agent}
\end{subfigure}
\caption{Excerpted prompt templates of \textsc{CoMIC}.}
\label{fig:prompt_templates}
\end{figure*}

\subsection{LLM-based Cloud Critic Framework}\label{sec:cloud}
The cloud critic operates asynchronously outside the immediate execution loop of the current edge episode, directly processing the uploaded interaction trajectories into structured reflections and reusable knowledge to guide future decisions at the edge.

\subsubsection{Single-Trajectory Canonicalization and Evaluation}
Cloud evaluation operates on individual subgoal trajectories independently. Uploaded trajectories are buffered in short-term memory without blocking edge execution. The immediate input to the cloud is therefore a single trajectory $\hat{\tau}^i_t$ coupled with its contextual features (e.g., task and subgoal identifiers), which allow the cloud to identify the corresponding subgoal episode.

Before evaluation, the contextual features of each trajectory are canonicalized into unified semantic identifiers to align heterogeneous edge inputs. Rather than matching raw subgoal strings, the cloud constructs a task-scoped identifier from the task type, the task variant, the normalized subgoal description, and the available formal metadata. For PDDL tasks, the subgoal is normalized to its predicate structure when available (e.g., ``stack the red block on the blue block'' becomes \texttt{stack(red, blue)}); for text-interaction tasks, the identifier retains a task-scoped stage description. This makes identifiers comparable within a task while preventing identical phrases in different tasks from being treated as the same subgoal.

The cloud then serializes each episode trajectory with its normalized features into a textual record:
\begin{equation}\label{eq:serialize}
x^i_t=\mathrm{Serialize}\big(\hat{\tau}^i_t,\tilde{m}^i_t\big),
\end{equation}
where $\tilde{m}^i_t$ denotes the canonicalized feature set and $x^i_t$ is the serialized evaluation record.

The cloud critic performs trajectory-level evaluation on $x^i_t$. To avoid redundant LLM calls, exact duplicate uploads are identified through cache identity and not re-evaluated. When no exact match is found, related prior experiences and synthesized global insights enrich the evaluation context, enabling the critic to assess the trajectory with reference to accumulated knowledge. Figure~\ref{prompt:cloud_critic} shows an excerpt of the cloud critic prompt, condensing its system prompt and user template.

\subsubsection{Critic-Guided Experience Distillation}
Based on the evaluation of the trajectory, the cloud critic further acts as a gated reflection module. The evaluation output $C^i_t$ forms a reflective record comprising a summary, insights, suggestions, and a self-reported confidence score $s_{\mathrm{self}}$ from the LLM. To assess the actual reliability of this reflection, we define an admission score for each evaluated trajectory record:
\begin{equation}\label{eq:adm}
s_{\mathrm{adm}}(x^i_t)=\mathrm{clip}_{[0,1]}\big(s_{\mathrm{self}}-\lambda_{\mathrm{ctx}}\,\mathbb{I}_{\mathrm{retry}}-\lambda_{\mathrm{short}}\,\mathbb{I}_{\mathrm{short}}\big),
\end{equation}
where $\mathbb{I}_{\mathrm{retry}}$ indicates that the evaluation relied on a retry with retrieved context, and $\mathbb{I}_{\mathrm{short}}$ indicates that the trajectory is too short to provide sufficient evidence.

A trajectory is regarded as too short if it contains fewer than 3 steps or fewer than 6 non-empty action-observation pairs. We set $\lambda_{\mathrm{ctx}}=0.10$ and $\lambda_{\mathrm{short}}=0.10$ empirically, which discount evaluations that required a context-assisted retry and trajectories with insufficient evidence, respectively. Since weak edge agents rarely produce successful trajectories, these penalties are designed to retain as many informative experiences from weak agents as possible.

\subsubsection{Memory Organization}
Memory within the cloud is organized into two components: a short-term buffering layer and a cloud knowledge base.

\paragraph{Short-term Memory.}
The cloud maintains a short-term buffer for recently uploaded trajectories and intermediate evaluations:
\begin{equation}\label{eq:stm}
\mathcal{M}_{\mathrm{STM}}=\mathcal{T}_{\mathrm{STM}}\cup\mathcal{E}_{\mathrm{STM}},
\end{equation}
where $\mathcal{T}_{\mathrm{STM}}$ contains recently uploaded subgoal episodes $\hat{\tau}^i_t$ together with their basic metadata, and $\mathcal{E}_{\mathrm{STM}}$ contains their temporary evaluation records for a short duration.

\paragraph{Cloud Knowledge Base.}
The cloud knowledge base comprises experience memory and global guidance. For experience memory, the evaluation of each trajectory record is distilled into generalized experience units $\mathcal{M}_{\mathrm{exp}}=\{\langle O,A,R\rangle\}$, where each tuple records an observation, an action, and the outcome. Each unit $m$ inherits the admission score of the trajectory record $x(m)$ from which it is distilled, and the admitted experience set is defined as
\begin{equation}\label{eq:admit}
\mathcal{M}^{+}_{\mathrm{exp}}=\big\{m\in\mathcal{M}_{\mathrm{exp}}\,\big|\,s_{\mathrm{adm}}(x(m))\ge\gamma_{\mathrm{kb}}\big\},
\end{equation}
where $\gamma_{\mathrm{kb}}$ denotes the knowledge-base admission threshold.

The threshold $\gamma_{\mathrm{kb}}$ depends on the capability of the source agent: we set $\gamma_{\mathrm{kb}}=0.60$ for trajectories from stronger edge agents and $\gamma_{\mathrm{kb}}=0.75$ for those from weak edge agents. The lower threshold collects more high-quality experiences from strong agents, while the higher threshold prevents noisy reflections on weak-agent trajectories from degrading the knowledge base. In heterogeneous deployments where both types of agents coexist, this setting obtains informative trajectories from weak agents while acquiring higher-quality trajectories from strong agents. All parameter values are fixed across tasks and are not claimed to be globally optimal.

These experiences form the foundation for experience aggregation in the cloud. For a grouping identifier $u$ of a given subgoal episode, the system extracts the corresponding experience group by indexing the admitted experience set:
\begin{equation}\label{eq:group}
\mathcal{G}_{\mathrm{exp}}(u)=\big\{m\in\mathcal{M}^{+}_{\mathrm{exp}}\,\big|\,\kappa(m)=u\big\},
\end{equation}
where $\kappa(m)$ denotes the grouping identity associated with experience item $m$. The retrieved group $\mathcal{G}_{\mathrm{exp}}(u)$ constitutes the historical experiences used to enrich the evaluation context and forms the basis for subsequent aggregation.

The guidance store is defined as
\begin{equation}\label{eq:store}
\mathcal{M}=\mathcal{M}_{\mathrm{manual}}\cup\mathcal{M}_{\mathrm{global}},
\end{equation}
where $\mathcal{M}_{\mathrm{manual}}$ represents predefined general task rules and $\mathcal{M}_{\mathrm{global}}$ represents reusable guidance generated by aggregating the experience groups above.

Because admitted experiences are indexed by subgoal group and duplicates are filtered by cache identity, retrieval is restricted to the matching group, so its cost does not grow with the number of edge agents. Each scenario in our experiments starts from an empty knowledge base, preventing experience carryover.

\subsection{Cloud-Edge Collaboration Mechanism}\label{sec:collab}
The cloud periodically aggregates admitted experiences to synthesize global guidance:
\begin{equation}\label{eq:agg}
G_{\mathrm{guid}}(u)=f_{\mathrm{agg}}\big(\mathcal{G}_{\mathrm{exp}}(u)\big).
\end{equation}

Grouping follows a three-level policy: experiences are first matched by exact identifier; when no exact match is available, the cloud falls back to task-level alignment within the same task setting; no free-text matching is performed across tasks. Coarse matching thus serves as a constrained fallback for incomplete or non-identical subgoal descriptions no as a general semantic-similarity search.

The synthesized guidance $G_{\mathrm{guid}}(u)$ comprises a reusable execution summary, empirical insights, prospective suggestions, and an aggregated credibility score. It is committed to $\mathcal{M}_{\mathrm{global}}$ and dispatched to edge agents addressing matching subgoals.

\subsubsection{Collaborative Context Assembly on Edge Agent}
Upon receiving global guidance from the cloud, the edge agent renders it into the unique \emph{Global Guidance} prompt block (Figure~\ref{prompt:edge_agent}). When initiating a new subgoal episode $t+1$, the agent employs a prompt assembly function $\mathcal{A}$ to construct the decision prompt $P^i_{t+1}$ from the global task $g$, the current subgoal $g^i_{t+1}$, the immediate observation $o^i_{t+1}$, the reconstructed history $H^i_{t+1}(\mathcal{I}_{t+1})$, and the guidance selected for the current identifier $u_{t+1}$:
\begin{equation}\label{eq:prompt}
P^i_{t+1}=\mathcal{A}\big(g,g^i_{t+1},o^i_{t+1},H^i_{t+1}(\mathcal{I}_{t+1}),G_{\mathrm{sel}}(u_{t+1})\big),
\end{equation}
where $G_{\mathrm{sel}}(u_{t+1})$ is selected from the guidance store $\mathcal{M}$ and is empty when no guidance is needed.

Trajectory admission runs after completed subgoal episodes are uploaded, and global guidance is synthesized in the cloud background. Runtime retrieval is invoked by the edge only when the current subgoal remains active and recent execution indicates that guidance is needed, and it returns guidance that has already been synthesized. The latest observation remains the authoritative environment state during action selection: when the guidance conflicts with the observed state, the agent can disregard it. Together with the admission penalties in Eq.~(\ref{eq:adm}), this mitigates the risk of unreliable reflections being repeatedly reinforced. The runtime procedure is summarized in Algorithm~\ref{alg:runtime-guidance-lifecycle} in Appendix~\ref{alg:runtime-guidance-procedure}.

\section{Experiments}
\subsection{Experimental Setup}\label{sec:setup}
\paragraph{\textbf{Environment.}}
All experiments are conducted without fine-tuning any cloud or edge LLM. Each edge agent runs on a Linux workstation with a single NVIDIA RTX 4090 GPU and communicates with the cloud through the Web gateway (\S\ref{sec:overview}); detailed runtime environments are provided in Appendix~\ref{app:runtime-env}.

\paragraph{\textbf{Evaluation Tasks.}}
We evaluate \textsc{CoMIC} on five AgentBoard benchmarks~\cite{Chang2024}, typically requiring over 20 decision steps: \textbf{Jericho}, which involves partially observable interaction through free form textual commands, and four symbolic planning domains, \textbf{Blocksworld}, \textbf{Gripper}, \textbf{Tyreworld}, and \textbf{Barman}, with strict action preconditions and long dependency chains. Blocksworld and Tyreworld contain 10 instances each; the others contain 20 each, spanning diverse initial states, object counts, and goals. Together, these benchmarks capture the long chains of dependent actions and textual interfaces characteristic of tasks served through the Web while avoiding the volatility of live websites.

\paragraph{\textbf{Deployment Scenarios.}}
We employ \texttt{deepseek-chat} as the cloud critic and design two edge scenarios. \textit{(i)} \textbf{Scenario A} (homogeneous lightweight edge) deploys only \texttt{Mistral-7b}~\cite{chaplot2023} agents. \textit{(ii)} \textbf{Scenario B} (heterogeneous mixed edge) deploys \texttt{Mistral-7b} agents together with a stronger \texttt{gpt-4-turbo}~\cite{Openai2023} agent.

\paragraph{\textbf{Baselines.}}
We compare against two baselines using the same \texttt{Mistral-7b} backbone. \textit{(i)} \textbf{\textsc{Standard}}, adapted from AgentBoard~\cite{Chang2024}, is a generic memory-based agent without hierarchical memory or subgoal decomposition. \textit{(ii)} \textbf{\textsc{Local}} builds upon HiAgent~\cite{Hu2025Hiagent}, a framework for long-horizon tasks with subtask-partitioned working memory; it corresponds to the edge agent of \textsc{CoMIC} without any cloud-side component. 

We do not adopt ExpeL~\cite{Zhao2024Expel} or Reflexion~\cite{Shinn2023} as baselines, since ExpeL extracts insights by comparing multiple successful trials, which weak edge agents rarely produce, and Reflexion accumulates verbal reflections over repeated trials, conflicting with the bounded context and latency budget of edge agents.

\paragraph{\textbf{Metrics.}}
\textbf{(1) Edge-side metrics} include Success Rate (\textbf{SR}), Progress Rate (\textbf{PR}), Grounding Accuracy (\textbf{GA}), \textbf{Steps}, and \textbf{Context}. SR is the proportion of task instances that are successfully completed, where an instance counts as completed only when all of its subgoals are achieved within the step limit ($H_{\max}=30$), and thus reflects the degree of task completion. PR is the proportion of completed subgoals in each instance, and thus reflects subgoal-level progress toward the final goal. Both are averaged over all instances of a task. GA measures the proportion of valid executed actions, and Steps counts the average number of decision steps taken per instance. Following HiAgent~\cite{Hu2025Hiagent}, Context counts the prompt tokens fed to the LLM at each decision step, tokenized with \texttt{cl100k\_base}; we average over steps within each instance and then over runs and instances. It is reported in relative form, $\mathrm{Context}(\%)=100\times\overline{T}_{\mathrm{compared}}/\overline{T}_{\mathrm{baseline}}$, where the baseline is \textsc{Standard} in Table~\ref{tab:scenario-a-analysis}, Scenario~A in Table~\ref{tab:scenario-b-analysis}, and \textsc{Local} in Table~\ref{tab:ablation-main}. The other rows in each table report absolute values together with their changes relative to this baseline.

% \textbf{\textit{(ii) Cloud-side metrics}} include the cache hit rate (\textbf{Hit}) of evaluation requests, response latency (\textbf{Lat.}), and the acknowledgement rate (\textbf{ACK}) of dispatched guidance; detailed definitions are provided in Appendix~\ref{app:metrics}. 

% \textbf{\textit{(iii) Protocol.}} The edge memory budget is $B=100$ items, and each task instance is limited to $H_{\max}=30$ decision steps. Each instance is run at least three times under a fixed model and inference configuration, and the results are consistent across runs.

\textbf{(2) Cloud-side metrics}, including the cache hit rate (\textbf{Hit}), response latency (\textbf{Lat.}), and acknowledgement rate (\textbf{ACK}), as well as the full evaluation protocol, are detailed in Appendix~\ref{app:metrics}.

\subsection{Main Results}

\begin{table*}[t]
\caption{Edge-side results in Scenario A.}
\label{tab:scenario-a-analysis}
\centering
\renewcommand{\arraystretch}{0.92}
\setlength{\tabcolsep}{10pt}
\begin{tabular}{lrlrlrlrlrl}
\toprule
\textbf{} & \textbf{SR $\uparrow$} &  & \textbf{PR $\uparrow$} &  & \textbf{Steps $\downarrow$} &  & \textbf{Context $\downarrow$} &  & \textbf{GA $\uparrow$} &  \\
\midrule
\textbf{\textit{Blocksworld}} & & & & & & & & & & \\
\textsc{Standard} & 0.00 &  & 5.00 &  & 30.00 &  & 100.00\% &  & 7.00 &  \\
\textsc{CoMIC} & \textbf{20.00} & {\footnotesize\color{green!50!black} +20.00} & \textbf{32.22} & {\footnotesize\color{green!50!black} +27.22} & \textbf{27.98} & {\footnotesize\color{green!50!black} -2.02} & \textbf{68.23\%} & {\footnotesize\color{green!50!black} -31.77\%} & \textbf{100.00} & {\footnotesize\color{green!50!black} +93.00} \\
\midrule
\textbf{\textit{Gripper}} & & & & & & & & & & \\
\textsc{Standard} & \textbf{0.00} &  & \textbf{4.12} &  & \textbf{30.00} &  & 100.00\% &  & 8.67 &  \\
\textsc{CoMIC} & \textbf{0.00} & {\footnotesize\color{green!50!black} +0.00} & 2.94 & {\footnotesize\color{red!70!black} -1.18} & \textbf{30.00} & {\footnotesize\color{green!50!black} +0.00} & \textbf{59.53\%} & {\footnotesize\color{green!50!black} -40.47\%} & \textbf{100.00} & {\footnotesize\color{green!50!black} +91.33} \\
\midrule
\textbf{\textit{Tyreworld}} & & & & & & & & & & \\
\textsc{Standard} & 0.00 &  & 10.89 &  & 30.00 &  & 100.00\% &  & 20.00 &  \\
\textsc{CoMIC} & \textbf{10.00} & {\footnotesize\color{green!50!black} +10.00} & \textbf{32.11} & {\footnotesize\color{green!50!black} +21.22} & \textbf{27.62} & {\footnotesize\color{green!50!black} -2.38} & \textbf{86.38\%} & {\footnotesize\color{green!50!black} -13.62\%} & \textbf{46.14} & {\footnotesize\color{green!50!black} +26.14} \\
\midrule
\textbf{\textit{Barman}} & & & & & & & & & & \\
\textsc{Standard} & 0.00 &  & 0.00 &  & 30.00 &  & 100.00\% &  & \textbf{29.50} &  \\
\textsc{CoMIC} & \textbf{2.50} & {\footnotesize\color{green!50!black} +2.50} & \textbf{6.39} & {\footnotesize\color{green!50!black} +6.39} & \textbf{29.35} & {\footnotesize\color{green!50!black} -0.65} & \textbf{69.60\%} & {\footnotesize\color{green!50!black} -30.40\%} & 27.08 & {\footnotesize\color{red!70!black} -2.42} \\
\midrule
\textbf{\textit{Jericho}} & & & & & & & & & & \\
\textsc{Standard} & \textbf{0.00} &  & \textbf{9.35} &  & 30.00 &  & 100.00\% &  & \textbf{97.50} &  \\
\textsc{CoMIC} & \textbf{0.00} & {\footnotesize\color{green!50!black} +0.00} & 6.27 & {\footnotesize\color{red!70!black} -3.08} & \textbf{29.53} & {\footnotesize\color{green!50!black} -0.47} & \textbf{91.94\%} & {\footnotesize\color{green!50!black} -8.06\%} & 96.50 & {\footnotesize\color{red!70!black} -1.00} \\
\hline
\midrule
\textbf{\textit{Overall}} & & & & & & & & & & \\
\textsc{Standard} & 0.00 &  & 5.87 &  & 30.00 &  & 100.00\% &  & 32.53 &  \\
\textsc{CoMIC} & \textbf{6.50} & {\footnotesize\color{green!50!black} +6.50} & \textbf{15.99} & {\footnotesize\color{green!50!black} +10.12} & \textbf{28.90} & {\footnotesize\color{green!50!black} -1.10} & \textbf{72.12\%} & {\footnotesize\color{green!50!black} -27.88\%} & \textbf{73.94} & {\footnotesize\color{green!50!black} +41.41} \\
\bottomrule
\end{tabular}
\renewcommand{\arraystretch}{1.0}
\end{table*}

% \begin{figure}[t]
%   \centering
%   \captionsetup[subfigure]{labelformat=simple}
%   \begin{subfigure}[t]{0.49\linewidth}
%     \centering
%     \includegraphics[width=\linewidth]{experiment/Scenario_A_pr_vs_context_fig_03.pdf}
%     \caption{Scenario A.}
%     \label{fig:scenario_a_pr_context_subfig}
%   \end{subfigure}
%   \hfill
%   \begin{subfigure}[t]{0.49\linewidth}
%     \centering
%     \includegraphics[width=\linewidth]{experiment/Scenario_B_pr_vs_context_fig_07.pdf}
%     \caption{Scenario B.}
%     \label{fig:scenario_b_pr_context_subfig}
%   \end{subfigure}
%   \caption{\textbf{Progress Rate vs. Context Token Consumption.} The plots show context token consumption across datasets against the corresponding progress rates, where asterisks denote averages over all environments. Scenario A shows that \textsc{CoMIC} improves task progression while reducing context cost relative to the \textsc{Standard} baseline, and Scenario B shows that trajectories generated by the stronger edge agent further improve progress on most tasks while reducing context consumption relative to Scenario A.}
%   \label{fig:scenario_a_pr_context}
% \end{figure}

\begin{figure*}[t]
  \centering
  \begin{minipage}[t]{0.69\textwidth}
    \centering
    \includegraphics[width=0.8\linewidth]{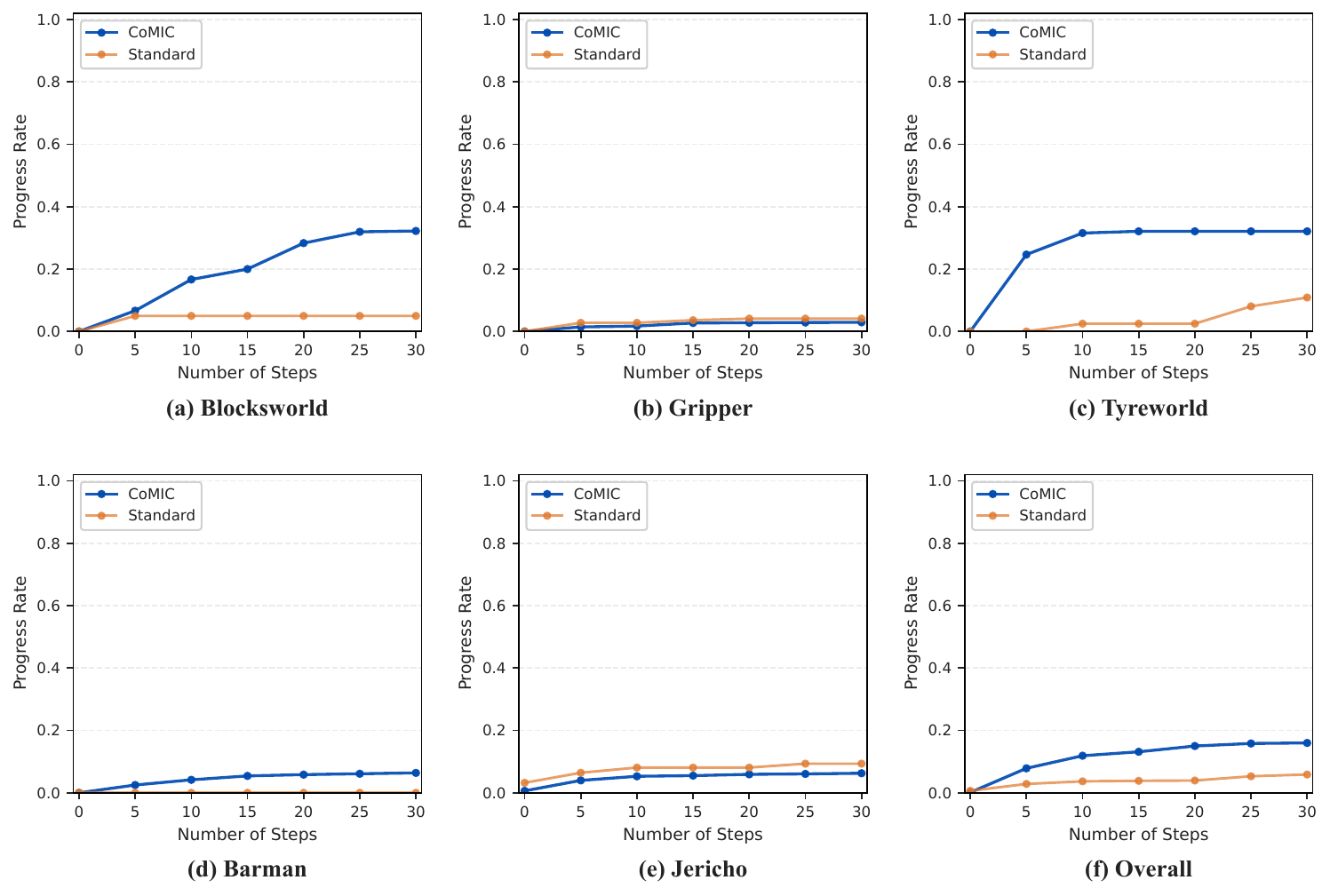}
    \caption{Progress rate vs.\ execution steps in Scenario A.}
    \label{fig:scenario_a_pr_steps}
  \end{minipage}
  \hfill
  \begin{minipage}[t]{0.29\textwidth}
    \centering
    \includegraphics[width=0.95\linewidth]{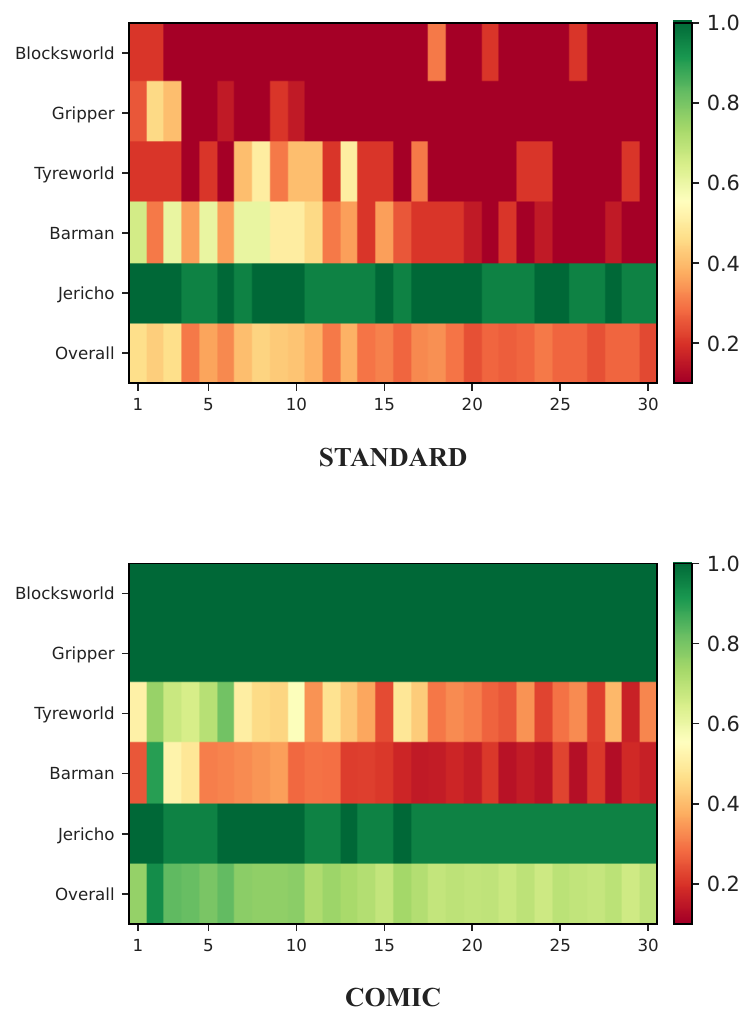}
    \caption{GA in Scenario A.}
    \label{fig:scenario_a_executability}
  \end{minipage}
\end{figure*}

\paragraph{\textbf{\textsc{CoMIC} brings weak edge agents from zero to non-zero success and improves their progress.}}
As shown in Table~\ref{tab:scenario-a-analysis}, the \textsc{Standard} agent exhausts the step budget on every task without completing a single instance, and most of its actions on the planning tasks are invalid. Under \textsc{CoMIC}, the same \texttt{Mistral-7b} agent completes instances on Blocksworld, Tyreworld, and Barman, where \textsc{Standard} fails entirely, and its overall PR nearly triples (5.87$\rightarrow$15.99). Grounding accuracy rises sharply on Blocksworld, Gripper, and Tyreworld (Figure~\ref{fig:scenario_a_executability}), indicating that the agent issues far fewer invalid actions once its decisions are organized around subgoals and informed by cloud guidance. These gains are achieved with fewer or equal steps on all tasks, and on Blocksworld, Tyreworld, and Barman the agent reaches higher progress within fewer steps (Figure~\ref{fig:scenario_a_pr_steps}). This indicates that even when trajectories come only from weak agents, the cloud critic can still extract reusable experience and return it as advisory guidance.

% \begin{figure}[t]
%   \centering
%   \includegraphics[width=\linewidth]{experiment/Scenario_A_pr_vs_steps_fig_01.pdf}
%   \caption{\textbf{Progress Rate vs. Execution Steps in Scenario A.} Compared to the standard baseline, the edge agent under the \textsc{CoMIC} framework achieves higher progress rates within the same or fewer execution steps across multiple environments, demonstrating enhanced action efficiency.}
%   \label{fig:scenario_a_pr_steps}
% \end{figure}

% \begin{figure}[t]
%   \centering
%   \begin{minipage}[t]{0.48\linewidth}
%     \centering
%     \includegraphics[width=\linewidth]{experiment/Scenario_A_executability_fig_02.pdf}
%     \caption{\textbf{Grounding Accuracy in Scenario A.} Assisted by cloud guidance, the edge agent issues fewer invalid actions on most tasks.}
%     \label{fig:scenario_a_executability}
%   \end{minipage}
%   \hfill
%   \begin{minipage}[t]{0.48\linewidth}
%     \centering
%     \includegraphics[width=\linewidth]{experiment/Scenario_B_executability_fig_06.pdf}
%     \caption{\textbf{Grounding Accuracy in Scenario B.} Guidance from stronger edge agents yields task-dependent gains.}
%     \label{fig:scenario_b_executability}
%   \end{minipage}
% \end{figure}

\paragraph{\textbf{\textsc{CoMIC} reduces context overhead.}}
Context consumption decreases on all five tasks, by 27.88\% overall (Table~\ref{tab:scenario-a-analysis}). The reduction holds even on Gripper and Jericho, where the success rate remains zero: although the weak agents do not yet produce enough high-quality experience for the cloud to drive task completion on these tasks, collaboration still keeps the decision prompts compact. Overall, \textsc{CoMIC} reaches higher progress at lower context cost on most tasks (Figure~\ref{fig:scenario_a_pr_context_subfig} in Appendix~\ref{app:pr-context}).

\paragraph{\textbf{Heterogeneous edge agents bring further gains.}}
A stronger edge agent in Scenario B further benefits the weak agent, improving SR, PR, and GA while reducing context consumption across tasks (7.73\% on average; Table~\ref{tab:scenario-b-analysis}). The weak agent achieves higher progress with lower context cost on most tasks (Figure~\ref{fig:scenario_b_pr_context_subfig} in Appendix~\ref{app:pr-context}), showing that stronger-peer guidance is more effective and concise.

\begin{table}[htbp]
\caption{Edge-side results in Scenario B.}
\label{tab:scenario-b-analysis}
\centering
\small
\renewcommand{\arraystretch}{0.92}
\setlength{\tabcolsep}{4pt}
\resizebox{\linewidth}{!}{%
\begin{tabular}{lrlrlrlrlrl}
\toprule
\textbf{} & \textbf{SR $\uparrow$} &  & \textbf{PR $\uparrow$} &  & \textbf{Steps $\downarrow$} &  & \textbf{Context $\downarrow$} &  & \textbf{GA $\uparrow$} &  \\
\midrule
\textbf{\textit{Blocksworld}} & & & & & & & & & & \\
\textsc{Scenario A} & \textbf{20.00} &  & 32.22 &  & 27.98 &  & 100.00\% &  & \textbf{100.00} &  \\
\textsc{Scenario B} & \textbf{20.00} & {\footnotesize\color{green!50!black} +0.00} & \textbf{33.33} & {\footnotesize\color{green!50!black} +1.11} & \textbf{27.80} & {\footnotesize\color{green!50!black} -0.18} & \textbf{96.10\%} & {\footnotesize\color{green!50!black} -3.90\%} & \textbf{100.00} & {\footnotesize\color{green!50!black} +0.00} \\
\midrule
\textbf{\textit{Gripper}} & & & & & & & & & & \\
\textsc{Scenario A} & \textbf{0.00} &  & 2.94 &  & \textbf{30.00} &  & 100.00\% &  & \textbf{100.00} &  \\
\textsc{Scenario B} & \textbf{0.00} & {\footnotesize\color{green!50!black} +0.00} & \textbf{6.42} & {\footnotesize\color{green!50!black} +3.47} & \textbf{30.00} & {\footnotesize\color{green!50!black} +0.00} & \textbf{97.75\%} & {\footnotesize\color{green!50!black} -2.25\%} & \textbf{100.00} & {\footnotesize\color{green!50!black} +0.00} \\
\midrule
\textbf{\textit{Tyreworld}} & & & & & & & & & & \\
\textsc{Scenario A} & 10.00 &  & 32.11 &  & 27.62 &  & 100.00\% &  & \textbf{46.14} &  \\
\textsc{Scenario B} & \textbf{12.50} & {\footnotesize\color{green!50!black} +2.50} & \textbf{32.50} & {\footnotesize\color{green!50!black} +0.39} & \textbf{27.25} & {\footnotesize\color{green!50!black} -0.37} & \textbf{91.89\%} & {\footnotesize\color{green!50!black} -8.11\%} & 41.69 & {\footnotesize\color{red!70!black} -4.45} \\
\midrule
\textbf{\textit{Barman}} & & & & & & & & & & \\
\textsc{Scenario A} & 2.50 &  & \textbf{6.39} &  & 29.35 &  & 100.00\% &  & 27.08 &  \\
\textsc{Scenario B} & \textbf{5.00} & {\footnotesize\color{green!50!black} +2.50} & 5.00 & {\footnotesize\color{red!70!black} -1.39} & \textbf{28.80} & {\footnotesize\color{green!50!black} -0.55} & \textbf{85.29\%} & {\footnotesize\color{green!50!black} -14.71\%} & \textbf{36.44} & {\footnotesize\color{green!50!black} +9.36} \\
\midrule
\textbf{\textit{Jericho}} & & & & & & & & & & \\
\textsc{Scenario A} & \textbf{0.00} &  & 6.27 &  & \textbf{29.53} &  & 100.00\% &  & 96.50 &  \\
\textsc{Scenario B} & \textbf{0.00} & {\footnotesize\color{green!50!black} +0.00} & \textbf{11.62} & {\footnotesize\color{green!50!black} +5.35} & 30.00 & {\footnotesize\color{red!70!black} +0.47} & \textbf{96.75\%} & {\footnotesize\color{green!50!black} -3.25\%} & \textbf{97.33} & {\footnotesize\color{green!50!black} +0.83} \\
\hline
\midrule
\textbf{\textit{Overall}} & & & & & & & & & & \\
\textsc{Scenario A} & 6.50 &  & 15.99 &  & 28.90 &  & 100.00\% &  & 73.94 &  \\
\textsc{Scenario B} & \textbf{7.50} & {\footnotesize\color{green!50!black} +1.00} & \textbf{17.77} & {\footnotesize\color{green!50!black} +1.79} & \textbf{28.77} & {\footnotesize\color{green!50!black} -0.13} & \textbf{92.27\%} & {\footnotesize\color{green!50!black} -7.73\%} & \textbf{75.09} & {\footnotesize\color{green!50!black} +1.15} \\
\bottomrule
\end{tabular}
}
\renewcommand{\arraystretch}{1.0}
\end{table}

\subsection{Analysis}
\subsubsection{Edge-side Analysis}\label{sec:hetero-analysis}
\begin{figure*}[t]
  \centering
  \begin{minipage}[t]{0.69\textwidth}
    \centering
    \includegraphics[width=0.8\linewidth]{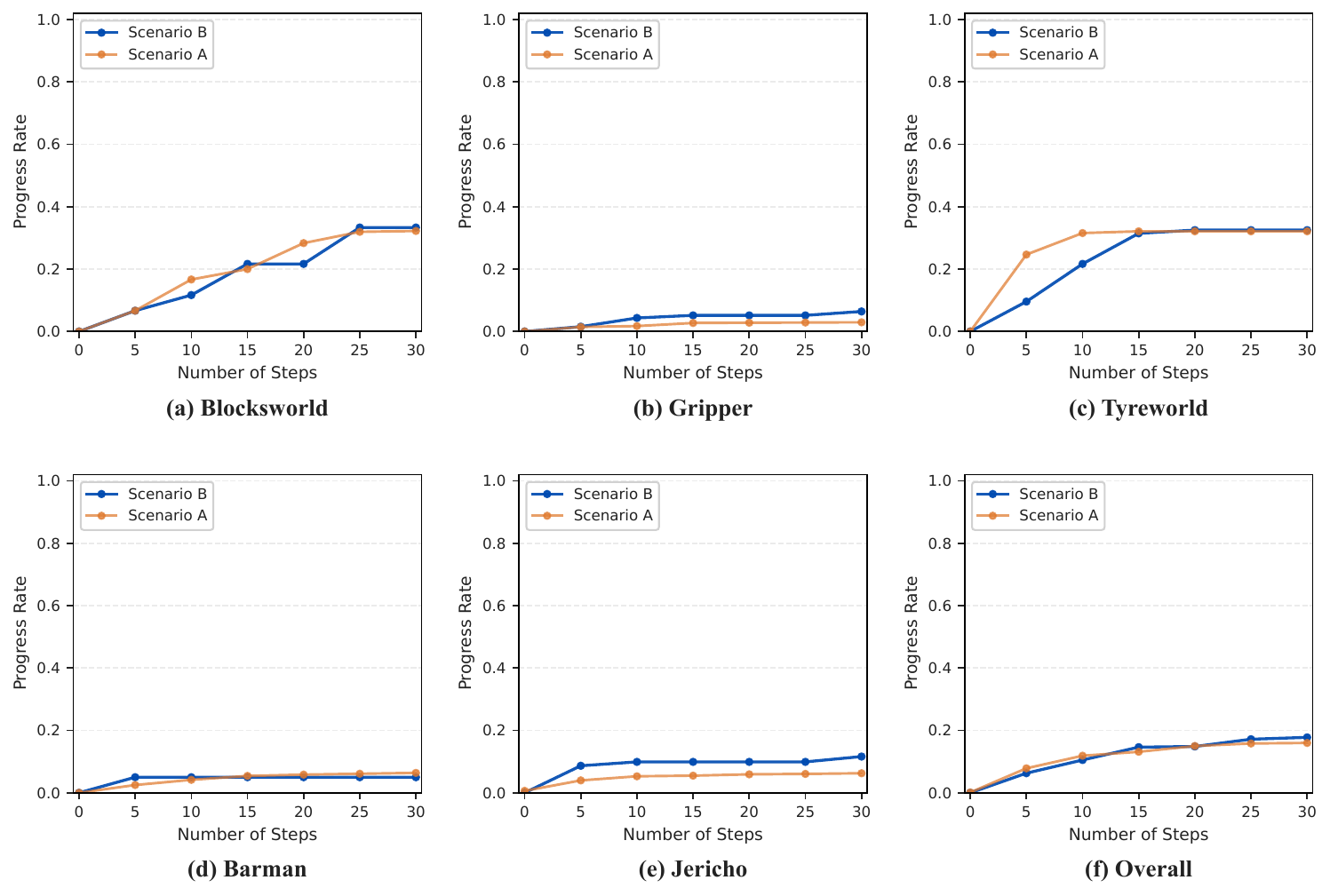}
    \caption{Progress rate vs.\ execution steps in Scenario B.}
    \label{fig:scenario_b_pr_steps}
  \end{minipage}
  \hfill
  \begin{minipage}[t]{0.29\textwidth}
    \centering
    \includegraphics[width=0.95\linewidth]{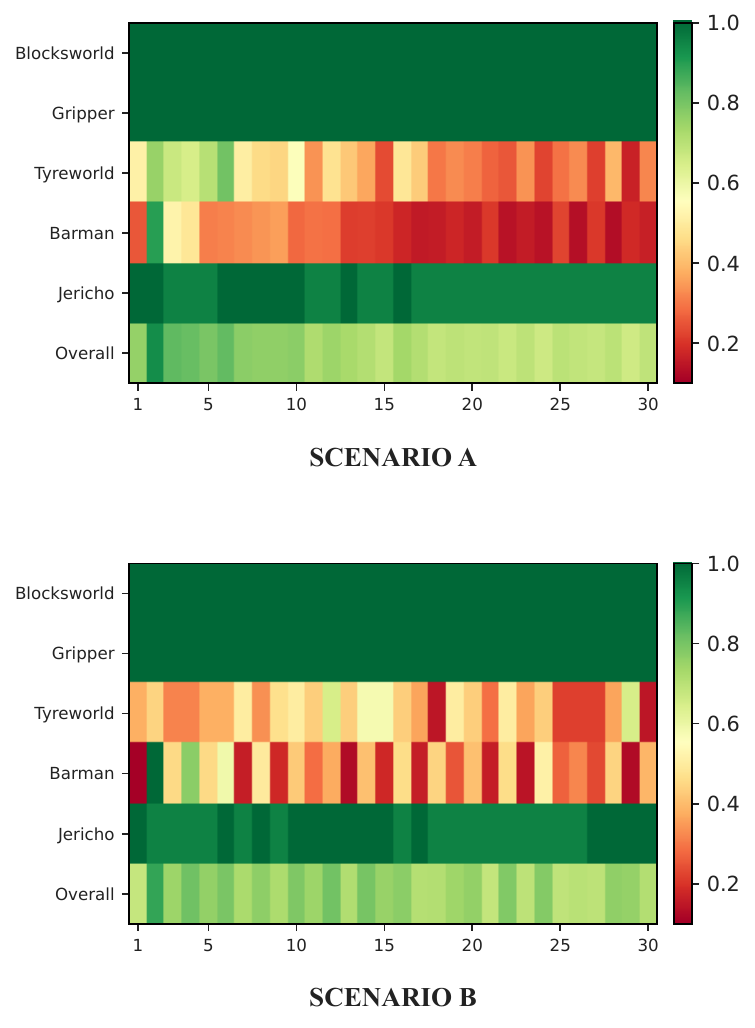}
    \caption{GA in Scenario B.}
    \label{fig:scenario_b_executability}
  \end{minipage}
\end{figure*}

Since the weak edge model and the cloud critic are identical in Scenarios A and B, the additional gains in Scenario B stem from cross-agent experience sharing: trajectories from the stronger agent provide the critic with more successful evidence, from which it distills more effective guidance for the weak agent. This effect is most visible on Gripper and Jericho: when the weak agent can only learn from its weak peers in Scenario A, its PR is even lower than that of \textsc{Local}, whereas in Scenario B its PR exceeds that of \textsc{Local} (Table~\ref{tab:ablation-main}; \S\ref{sec:ablation}). As shown in Figure~\ref{fig:scenario_b_pr_steps}, these progress gains are achieved with execution lengths comparable to those in Scenario A. Success on these tasks, however, remains bounded by the weak base model: since it struggles to plan a correct sequence of subgoals, its subgoals often fail to match those under which global guidance is indexed, preventing it from retrieving the corresponding guidance. Nevertheless, within reasonably planned subgoal episodes, the weak agent in Scenario B receives more effective guidance through heterogeneous collaboration, and its overall grounding accuracy is higher than that in Scenario A (Figure~\ref{fig:scenario_b_executability}).

% \begin{figure}[t]
%   \centering
%   \includegraphics[width=\linewidth]{experiment/Scenario_B_pr_vs_steps_fig_05.pdf}
%   \caption{\textbf{Progress Rate vs. Execution Steps in Scenario B.} Compared to Scenario A, the weak edge in Scenario B shows selective progress gains while maintaining comparable execution length across environments.}
%   \label{fig:scenario_b_pr_steps}
% \end{figure}

\subsubsection{Cloud-side Analysis}
In Scenario A, the cloud achieves an average Hit rate of 45.12\%, a response latency of 68.68\,s, and an ACK rate of 71.00\%. Although the latency is substantial, it does not delay edge execution, since trajectory uploading and guidance synthesis proceed asynchronously and the number of steps does not increase. The ACK rate shows that most dispatched guidance is acknowledged by the edge agents, while the remainder is disregarded when it conflicts with the current observation. In Scenario B, Hit and ACK decrease to 23.47\% and 40.00\%, as guidance distilled from the stronger agent is more abstract and thus harder for the weak agent to apply; nevertheless, the weak agent achieves better edge-side performance when it does apply such guidance. Detailed coordination statistics and the cloud pipeline, guidance, and overhead metrics are provided in Appendices~\ref{app:cloud-metrics} and~\ref{app:collab-metrics}, respectively.

\subsubsection{Ablation Study}\label{sec:ablation}
% Table~\ref{tab:ablation-main} compares three configurations with the same weak \texttt{Mistral-7b} executor: \textsc{Local} (\emph{w/o Cloud}), Scenario A (\emph{w/ Cloud}), and Scenario B (\emph{w/ Hetero.\ Cloud}); task-level results are provided in Appendix~\ref{app:ablation-details}. In Scenario A, cloud collaboration raises SR from 3.00 to 6.50 and PR from 12.35 to 15.99 with fewer steps, while GA slightly decreases and context consumption increases by 6.64\%. This comparison reflects that, on top of the hierarchical memory at the edge, cloud collaboration mainly contributes to task completion and progress. The changes in GA and context arise from the scarcity of informative trajectories in Scenario A: since all edge agents are weak and most of their trajectories are unsuccessful, the cloud critic summarizes guidance from noisy evidence, which yields longer guidance blocks and leads the agent to attempt a wider range of actions rather than repeating a narrow set of safe ones. Although some of these attempts are invalid, they allow the agent to complete more subgoals than under purely local execution.

% \input{table/t2_ablation.tex}

Table~\ref{tab:ablation-main} compares three configurations with the same weak \texttt{Mistral-7b} executor: \textsc{Local} (\emph{w/o Cloud}), Scenario A (\emph{w/ Cloud}), and Scenario B (\emph{w/ Hetero.\ Cloud}). Without the cloud, \textsc{Local} achieves low SR and PR on nearly all tasks and often exhausts the step budget. In Scenario A, cloud collaboration raises the overall SR from 3.00 to 6.50 and PR from 12.35 to 15.99 with fewer steps, while GA slightly decreases and context consumption increases by 6.64\%. This comparison reflects that, on top of the hierarchical memory at the edge, cloud collaboration mainly contributes to task completion and progress. The changes in GA and context arise from the scarcity of informative trajectories in Scenario A: since all edge agents are weak and most of their trajectories are unsuccessful, the cloud critic summarizes guidance from noisy evidence, which yields longer guidance blocks and leads the agent to attempt a wider range of actions rather than repeating a narrow set of safe ones. Although some of these attempts are invalid, they allow the agent to complete more subgoals than under purely local execution, and in long-horizon tasks, each completed subgoal episode represents concrete progress. In Scenario B, the stronger agent provides more successful trajectories, and the weak agent further raises the overall SR and PR to 7.50 and 17.77 while consuming less context than \textsc{Local}, suggesting that the overhead in Scenario A stems from the scarcity of informative trajectories rather than from cloud collaboration itself.

\begin{table}[htbp]
\caption{Task-level ablation results.}
% \label{tab:ablation-details}
 \label{tab:ablation-main}
\centering
\small
\renewcommand{\arraystretch}{0.92}
\setlength{\tabcolsep}{4pt}
\resizebox{\linewidth}{!}{%
\begin{tabular}{lrlrlrlrlrl}
\toprule
\textbf{} & \textbf{SR $\uparrow$} &  & \textbf{PR $\uparrow$} &  & \textbf{Steps $\downarrow$} &  & \textbf{Context $\downarrow$} &  & \textbf{GA $\uparrow$} &  \\
\midrule
\textbf{\textit{Blocksworld}} & & & & & & & & & & \\
\textsc{Local} & 0.00 & & 6.67 & & 30.00 & & \textbf{100.00\%} & & \textbf{100.00} & \\
\textsc{Scenario A} & \textbf{20.00} & {\footnotesize\color{green!50!black} +20.00} & 32.22 & {\footnotesize\color{green!50!black} +25.56} & 27.98 & {\footnotesize\color{green!50!black} -2.02} & 106.84\% & {\footnotesize\color{red!70!black} +6.84\%} & \textbf{100.00} & {\footnotesize\color{green!50!black} +0.00} \\
\textsc{Scenario B} & \textbf{20.00} & {\footnotesize\color{green!50!black} +20.00} & \textbf{33.33} & {\footnotesize\color{green!50!black} +26.67} & \textbf{27.80} & {\footnotesize\color{green!50!black} -2.20} & 102.68\% & {\footnotesize\color{red!70!black} +2.68\%} & \textbf{100.00} & {\footnotesize\color{green!50!black} +0.00} \\
\midrule
\textbf{\textit{Gripper}} & & & & & & & & & & \\
\textsc{Local} & \textbf{0.00} & & 3.50 & & \textbf{30.00} & & \textbf{100.00\%} & & \textbf{100.00} & \\
\textsc{Scenario A} & \textbf{0.00} & {\footnotesize\color{green!50!black} +0.00} & 2.94 & {\footnotesize\color{red!70!black} -0.56} & \textbf{30.00} & {\footnotesize\color{green!50!black} +0.00} & 102.81\% & {\footnotesize\color{red!70!black} +2.81\%} & \textbf{100.00} & {\footnotesize\color{green!50!black} +0.00} \\
\textsc{Scenario B} & \textbf{0.00} & {\footnotesize\color{green!50!black} +0.00} & \textbf{6.42} & {\footnotesize\color{green!50!black} +2.92} & \textbf{30.00} & {\footnotesize\color{green!50!black} +0.00} & 100.50\% & {\footnotesize\color{red!70!black} +0.50\%} & \textbf{100.00} & {\footnotesize\color{green!50!black} +0.00} \\
\midrule
\textbf{\textit{Tyreworld}} & & & & & & & & & & \\
\textsc{Local} & 10.00 & & \textbf{37.67} & & 29.20 & & 100.00\% & & \textbf{48.30} & \\
\textsc{Scenario A} & 10.00 & {\footnotesize\color{green!50!black} +0.00} & 32.11 & {\footnotesize\color{red!70!black} -5.56} & 27.62 & {\footnotesize\color{green!50!black} -1.58} & 100.51\% & {\footnotesize\color{red!70!black} +0.51\%} & 46.14 & {\footnotesize\color{red!70!black} -2.16} \\
\textsc{Scenario B} & \textbf{12.50} & {\footnotesize\color{green!50!black} +2.50} & 32.50 & {\footnotesize\color{red!70!black} -5.17} & \textbf{27.25} & {\footnotesize\color{green!50!black} -1.95} & \textbf{92.36\%} & {\footnotesize\color{green!50!black} -7.64\%} & 41.69 & {\footnotesize\color{red!70!black} -6.62} \\
\midrule
\textbf{\textit{Barman}} & & & & & & & & & & \\
\textsc{Local} & \textbf{5.00} & & 5.00 & & \textbf{28.75} & & \textbf{100.00\%} & & \textbf{37.67} & \\
\textsc{Scenario A} & 2.50 & {\footnotesize\color{red!70!black} -2.50} & \textbf{6.39} & {\footnotesize\color{green!50!black} +1.39} & 29.35 & {\footnotesize\color{red!70!black} +0.60} & 118.62\% & {\footnotesize\color{red!70!black} +18.62\%} & 27.08 & {\footnotesize\color{red!70!black} -10.58} \\
\textsc{Scenario B} & \textbf{5.00} & {\footnotesize\color{green!50!black} +0.00} & 5.00 & {\footnotesize\color{green!50!black} +0.00} & 28.80 & {\footnotesize\color{red!70!black} +0.05} & 101.18\% & {\footnotesize\color{red!70!black} +1.18\%} & 36.44 & {\footnotesize\color{red!70!black} -1.22} \\
\midrule
\textbf{\textit{Jericho}} & & & & & & & & & & \\
\textsc{Local} & \textbf{0.00} & & 8.91 & & \textbf{29.50} & & 100.00\% & & 97.00 & \\
\textsc{Scenario A} & \textbf{0.00} & {\footnotesize\color{green!50!black} +0.00} & 6.27 & {\footnotesize\color{red!70!black} -2.64} & 29.53 & {\footnotesize\color{red!70!black} +0.03} & 98.84\% & {\footnotesize\color{green!50!black} -1.16\%} & 96.50 & {\footnotesize\color{red!70!black} -0.50} \\
\textsc{Scenario B} & \textbf{0.00} & {\footnotesize\color{green!50!black} +0.00} & \textbf{11.62} & {\footnotesize\color{green!50!black} +2.72} & 30.00 & {\footnotesize\color{red!70!black} +0.50} & \textbf{95.63\%} & {\footnotesize\color{green!50!black} -4.37\%} & \textbf{97.33} & {\footnotesize\color{green!50!black} +0.33} \\
\hline
\midrule
\textbf{\textit{Overall}} & & & & & & & & & & \\
\textsc{Local} & 3.00 & & 12.35 & & 29.49 & & 100.00\% & & \textbf{76.59} & \\
\textsc{Scenario A} & 6.50 & {\footnotesize\color{green!50!black} +3.50} & 15.99 & {\footnotesize\color{green!50!black} +3.64} & 28.90 & {\footnotesize\color{green!50!black} -0.59} & 106.64\% & {\footnotesize\color{red!70!black} +6.64\%} & 73.94 & {\footnotesize\color{red!70!black} -2.65} \\
\textsc{Scenario B} & \textbf{7.50} & {\footnotesize\color{green!50!black} +4.50} & \textbf{17.77} & {\footnotesize\color{green!50!black} +5.43} & \textbf{28.77} & {\footnotesize\color{green!50!black} -0.72} & \textbf{98.40\%} & {\footnotesize\color{green!50!black} -1.60\%} & 75.09 & {\footnotesize\color{red!70!black} -1.50} \\
\bottomrule
\end{tabular}
}
\renewcommand{\arraystretch}{1.0}
\end{table}

\section{Limitations}
\label{sec:limitations}
While \textsc{CoMIC} improves the execution capabilities of weak edge agents, its end-to-end task success remains constrained by the planning ability of the edge base model. An excessively weak backbone struggles to generate viable subgoals or fully comprehend the global guidance from the cloud critic, and the subgoals it plans may fail to match those under which global guidance is indexed. Moreover, the benefit of cloud-edge collaboration is bounded by the informativeness of the trajectories produced by edge agents. In homogeneous weak-edge settings such as Scenario~A, most trajectories are unsuccessful and contribute limited reusable knowledge, which makes it difficult to maintain a high-quality knowledge base and, as shown in the ablation study, can increase context consumption relative to purely local execution.

In addition, our evaluation uses AgentBoard benchmarks that share the characteristics of Web-served long-horizon tasks and deploys a small number of edge agents in each scenario; the behavior of \textsc{CoMIC} on live websites, with larger edge deployments, and under realistic network conditions remains to be studied.

\section{Conclusion}
In this paper, we propose \textsc{CoMIC}, a parameter-update-free cloud-edge collaborative framework for the long-horizon decision-making of lightweight LLM agents deployed on Web edge servers. Following the paradigm of ``Act at the Edge, Reflect in the Cloud'', edge agents execute subgoals with hierarchical memory, while a cloud critic evaluates the trajectories uploaded by all edge agents and circulates the distilled global guidance back to them. Experiments on five long-horizon benchmarks show that, when only weak edge agents are deployed, \textsc{CoMIC} raises their success rate from zero to non-zero on Blocksworld, Tyreworld, and Barman, where the \textsc{Standard} agent completes no instance, and improves their progress rate over both the \textsc{Standard} agent and purely local execution; when stronger heterogeneous agents join the edge, the weak agents further improve in success, progress, and grounding with lower context consumption.

Future work will evaluate \textsc{CoMIC} on live Web tasks and larger edge deployments, refine the granularity of cloud guidance, and integrate reinforcement learning to adapt the confidence thresholds for cloud-side reflection and the management of memory.

%% ---------- Acknowledgments (keep commented out for anonymous submission) ----------
% \begin{acks}
% \end{acks}

%% ---------- Ethics statement (optional, encouraged by WWW) ----------
% \section*{Ethics and Privacy Statement}

%% ---------- References ----------
\bibliographystyle{ACM-Reference-Format}
\bibliography{references}

%% ---------- Appendix (optional; references + appendix up to 12 pages total) ----------
\appendix
\section{Appendix}
% \subsection{More Details on Background}
% \label{sec:appendix-background}
% Figure~\ref{fig:cloud-edge-end-scenario} illustrates the detailed cloud-edge-end system architecture on the Web.

% \begin{figure}[htbp]
%   \centering
%   \includegraphics[width=\linewidth]{figures/Overview.pdf}
%   \caption{\textbf{Cloud-edge-end system.} Deploying lightweight LLM agents on Web edge servers enables localized services for end users. However, constrained by scale-out deployment costs, widely distributed edge nodes cannot host massive models with parameter scales comparable to cloud servers, which strictly limits the reasoning capabilities, memory capacities, and computational resources of edge agents. When facing the demands of complex long-horizon tasks, these resource-bounded and isolated edge agents are unable to extract sufficient prior experiences solely from their limited local memories, highlighting the critical gap between localized resource constraints and complex task requirements.}
%   \label{fig:cloud-edge-end-scenario}
% \end{figure}

\subsection{Runtime Guidance Update Procedure}
\label{alg:runtime-guidance-procedure}
% Algorithm~\ref{alg:runtime-guidance-lifecycle} summarizes the implementation-level lifecycle that connects cloud-side trajectory evaluation with edge-side runtime guidance. It is not a synchronous control loop executed at every decision step: trajectory admission runs after completed subgoal episodes are uploaded, global guidance is synthesized in the cloud background, and runtime retrieval is invoked by the edge only when the current subgoal remains active and recent execution indicates that guidance is needed. The latest observation remains the authoritative environment state during action selection.
Algorithm~\ref{alg:runtime-guidance-lifecycle} summarizes the implementation-level lifecycle described in \S\ref{sec:collab}, which connects cloud-side trajectory evaluation with edge-side runtime guidance.

\begin{algorithm}[t]
\caption{Asynchronous lifecycle of trajectory admission and runtime guidance}
\label{alg:runtime-guidance-lifecycle}
\begin{algorithmic}[1]
\REQUIRE Completed subgoal episode $\hat{\tau}_t^i$, metadata $\tilde{m}_t^i$, current identifier $u_{t+1}$, observation $o_{t+1}^i$, history indices $\mathcal{I}_{t+1}$, and guidance store $\mathcal{M}=\mathcal{M}_{\mathrm{manual}}\cup\mathcal{M}_{\mathrm{global}}$.
\ENSURE Updated guidance store $\mathcal{M}$ and next prompt $P_{t+1}^i$ with at most one \textit{Global Guidance} block.
\STATE \colorbox{gray!12}{\parbox{0.95\linewidth}{\textit{Cloud admission path}}}
\STATE \textbf{Serialize:} $x_t^i \leftarrow \mathrm{Serialize}(\hat{\tau}_t^i,\tilde{m}_t^i)$.
\IF{$x_t^i \in \mathrm{cache}$}
  \STATE \textbf{Reuse} the cached trajectory evaluation record.
\ELSE
  \STATE \textbf{Upload} $x_t^i$ to $\mathcal{M}_{\mathrm{STM}}$.
  \STATE \textbf{Evaluate} $x_t^i$ with the Cloud Critic and compute $s_{\mathrm{adm}}(x_t^i)$.
  \STATE \textbf{Distill} reusable experience units $m$ from $x_t^i$, each inheriting $s_{\mathrm{adm}}(x_t^i)$.
  \IF{$s_{\mathrm{adm}}(x_t^i)\ge \gamma_{\mathrm{kb}}$ and no hard block}
    \STATE \textbf{Admit} the units into $\mathcal{M}_{\mathrm{exp}}^{+}$. \hfill $\triangleright$ Eq.~\eqref{eq:admit}
  \ENDIF
\ENDIF
\STATE \colorbox{gray!12}{\parbox{0.95\linewidth}{\textit{Cloud aggregation path}}}
\FOR{each eligible exact or coarse group $u$}
  \STATE \textbf{Retrieve} $\mathcal{G}_{\mathrm{exp}}(u)\leftarrow\{m\in\mathcal{M}_{\mathrm{exp}}^{+}\mid\kappa(m)=u\}$.
  \STATE \textbf{Update} $\mathcal{M}_{\mathrm{global}}$ with $G_{\mathrm{guid}}(u)\leftarrow f_{\mathrm{agg}}(\mathcal{G}_{\mathrm{exp}}(u))$. \hfill $\triangleright$ Eq.~\eqref{eq:agg}
\ENDFOR
\STATE \colorbox{gray!12}{\parbox{0.95\linewidth}{\textit{Edge runtime path}}}
\IF{runtime guidance is triggered}
  \STATE \textbf{Select} $G_{\mathrm{sel}}(u_{t+1})$ from $\mathcal{M}$ using the SOP Selector.
  \STATE \textbf{Ground} any recommended step against $o_{t+1}^i$ and valid actions when provided.
\ELSE
  \STATE Set $G_{\mathrm{sel}}(u_{t+1})\leftarrow \varnothing$.
\ENDIF
\STATE Assemble $P_{t+1}^i\leftarrow \mathcal{A}(g,g_{t+1}^i,o_{t+1}^i,H_{t+1}^i(\mathcal{I}_{t+1}),G_{\mathrm{sel}}(u_{t+1}))$.
\STATE \textbf{Execute} with $o_{t+1}^i$ as the authoritative environment state.
\RETURN $\mathcal{M}$ and $P_{t+1}^i$.
\end{algorithmic}
\end{algorithm}

\subsection{Runtime Environments}
\label{app:runtime-env}
Table~\ref{tab:runtime-environments} summarizes the runtime environments used for different experimental components. The \textsc{CoMIC} edge agents and the two baselines run \texttt{Mistral-7b} locally with the same GPU and software stack, and differ only in the CPU model. Since the edge-side metrics are computed from task outcomes and token counts, this difference does not affect the comparison. The cloud server, which hosts the Web gateway and the cloud-side memory stores, runs on an Apple M2 machine, while the cloud critic (\texttt{deepseek-chat}) and the strong edge agent in Scenario~B (\texttt{gpt-4-turbo}) are accessed through their official APIs.

% \begin{table*}[t]
%   \centering
%   \caption{Runtime environments for different experimental components.}
%   \label{tab:runtime-environments}
%   \small
%   \begin{tabular}{lllll}
%     \toprule
%     \textbf{Component} & \textbf{GPU} & \textbf{CPU} & \textbf{Memory} & \textbf{Software} \\
%     \midrule
%     \textsc{CoMIC} Edge    & RTX 4090 (24 GB) & Intel Xeon Gold 6430     & 120 GB & Python 3.8, PyTorch 2.0.0, CUDA 11.8 \\
%     \textsc{Local}         & RTX 4090 (24 GB) & Intel Xeon Platinum 8358P & 120 GB & Python 3.8, PyTorch 2.0.0, CUDA 11.8 \\
%     \textsc{Standard}      & RTX 4090 (24 GB) & Intel Xeon Platinum 8358P & 120 GB & Python 3.8, PyTorch 2.0.0, CUDA 11.8 \\
%     \textsc{CoMIC} Cloud$^{\dagger}$ & --     & Apple M2                 & 8 GB   & macOS Tahoe 26.0.1 \\
%     \bottomrule
%   \end{tabular}
%   \par\smallskip
% \end{table*}

\begin{table}[htbp]
  \centering
  \caption{Runtime environments.}
  \label{tab:runtime-environments}
  \small
  \renewcommand{\arraystretch}{1.1}
  \setlength{\tabcolsep}{4pt}
  \newcommand{\cell}[1]{\begin{tabular}[t]{@{}l@{}}#1\end{tabular}}
  \resizebox{\linewidth}{!}{%
  \begin{tabular}{lllll}
    \toprule
    \textbf{Component} & \textbf{GPU} & \textbf{CPU} & \textbf{Memory} & \textbf{Software} \\
    \midrule
    \textsc{CoMIC} Edge & \cell{RTX 4090\\(24 GB)} & \cell{Intel Xeon\\Gold 6430}      & 120 GB & \cell{Python 3.8, PyTorch 2.0.0,\\CUDA 11.8} \\
    \textsc{Local}      & \cell{RTX 4090\\(24 GB)} & \cell{Intel Xeon\\Platinum 8358P} & 120 GB & \cell{Python 3.8, PyTorch 2.0.0,\\CUDA 11.8} \\
    \textsc{Standard}   & \cell{RTX 4090\\(24 GB)} & \cell{Intel Xeon\\Platinum 8358P} & 120 GB & \cell{Python 3.8, PyTorch 2.0.0,\\CUDA 11.8} \\
    \textsc{CoMIC} Cloud & --                      & Apple M2                          & 8 GB   & macOS Tahoe 26.0.1 \\
    \bottomrule
  \end{tabular}
  }
  \renewcommand{\arraystretch}{1.0}
\end{table}

\subsection{Detailed Evaluation Metrics and Protocol}
\label{app:metrics}
In addition to the edge-side metrics defined in \S\ref{sec:setup}, we report the following cloud-side and cloud-edge collaboration metrics.

% \textbf{Edge-side metrics} include: \textit{(i)} \textbf{Success Rate (SR)}, the proportion of task instances that are successfully completed, reflecting task completion; \textit{(ii)} \textbf{Progress Rate (PR)}, the proportion of completed subgoals, reflecting subgoal progress; \textit{(iii)} \textbf{Grounding Accuracy (GA)}, the proportion of valid executed actions in the environment; \textit{(iv)} \textbf{Steps}, the average number of execution steps required to finish or terminate a task; and \textit{(v)} \textbf{Context}, the scale of prompt tokens consumed during execution.

% Following HiAgent~\cite{Hu2025Hiagent}, Context counts the prompt tokens fed to the LLM at each decision step, tokenized with \texttt{cl100k\_base}. We first average over steps within each task instance, and then average over runs and instances. Context is reported in relative form, $\mathrm{Context}(\%)=100\times\overline{T}_{\mathrm{compared}}/\overline{T}_{\mathrm{baseline}}$, where $\overline{T}_{\mathrm{baseline}}$ is the average token count of the reference setting in each table, i.e., \textsc{Standard} in Table~\ref{tab:scenario-a-analysis}, Scenario~A in Table~\ref{tab:scenario-b-analysis}, and \textsc{Local} in the ablation.

\textbf{Cloud-side metrics} evaluate the cloud's response capability and delivery effectiveness. They include \textbf{Cache Hit Rate (Hit)}, the proportion of requests served directly from cached evaluation records; \textbf{Confidence Score (Conf.)}, the average self-reported confidence of the cloud critic; cloud response latency (\textbf{Lat.}), the average response time; \textbf{Insights Count (Ins.)}, the average number of generated insights; and \textbf{ACK Rate (ACK)}, the proportion of dispatched guidance acknowledged by edge agents.

\textbf{Cloud-edge collaboration metrics} evaluate the internal processing and resource consumption of the cloud memory driven by edge-cloud interaction, comprising two components: \textit{(i)} \textbf{Cloud Critic Metrics} evaluate the experience critic process. This category encompasses \textit{cloud pipeline metrics}--tracking the edge agent's experience \textbf{Upload}s, evaluated \textbf{Response}s, and admitted rules (\textbf{KB Admit})--and \textit{cloud guidance metrics}, which trace the guidance lifecycle through generated \textbf{Insight}s, acknowledged guidance (\textbf{ACK}), and successfully executed actions (\textbf{Adopt}); and \textit{(ii)} \textbf{Cloud Overhead Metrics} quantify the total cost of resource consumption (\textbf{Cloud Total}), which comprises the distillation pipeline cost (\textbf{Pipeline Total}) and the guidance retrieval cost (\textbf{Guidance Total}).

\textbf{Protocol.} The edge memory budget is $B=100$ items, and each task instance is limited to $H_{\max}=30$ decision steps. Each instance is run at least three times under a fixed model and inference configuration, and the results are consistent across runs.

\subsection{Additional Edge-side Analysis}
\label{app:pr-context}
\begin{figure}[t]
  \centering
  \captionsetup[subfigure]{labelformat=simple}
  \begin{subfigure}[t]{0.49\linewidth}
    \centering
    \includegraphics[width=\linewidth]{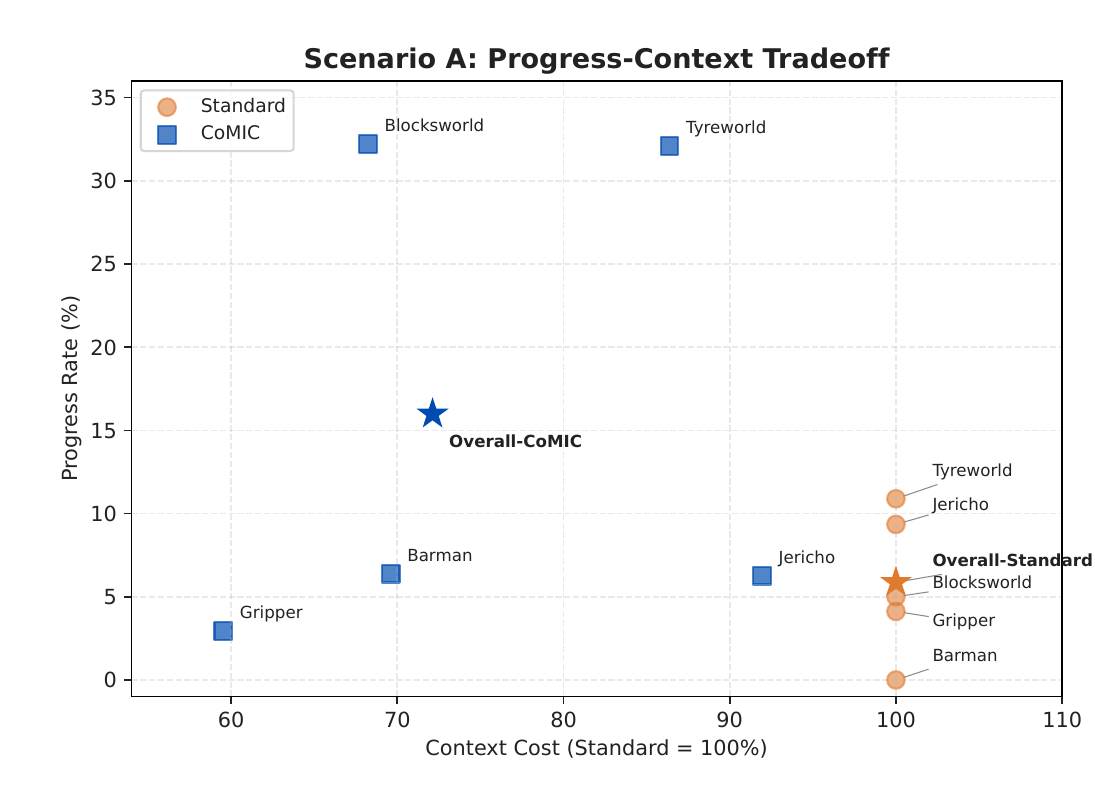}
    \caption{Scenario A.}
    \label{fig:scenario_a_pr_context_subfig}
  \end{subfigure}
  \hfill
  \begin{subfigure}[t]{0.49\linewidth}
    \centering
    \includegraphics[width=\linewidth]{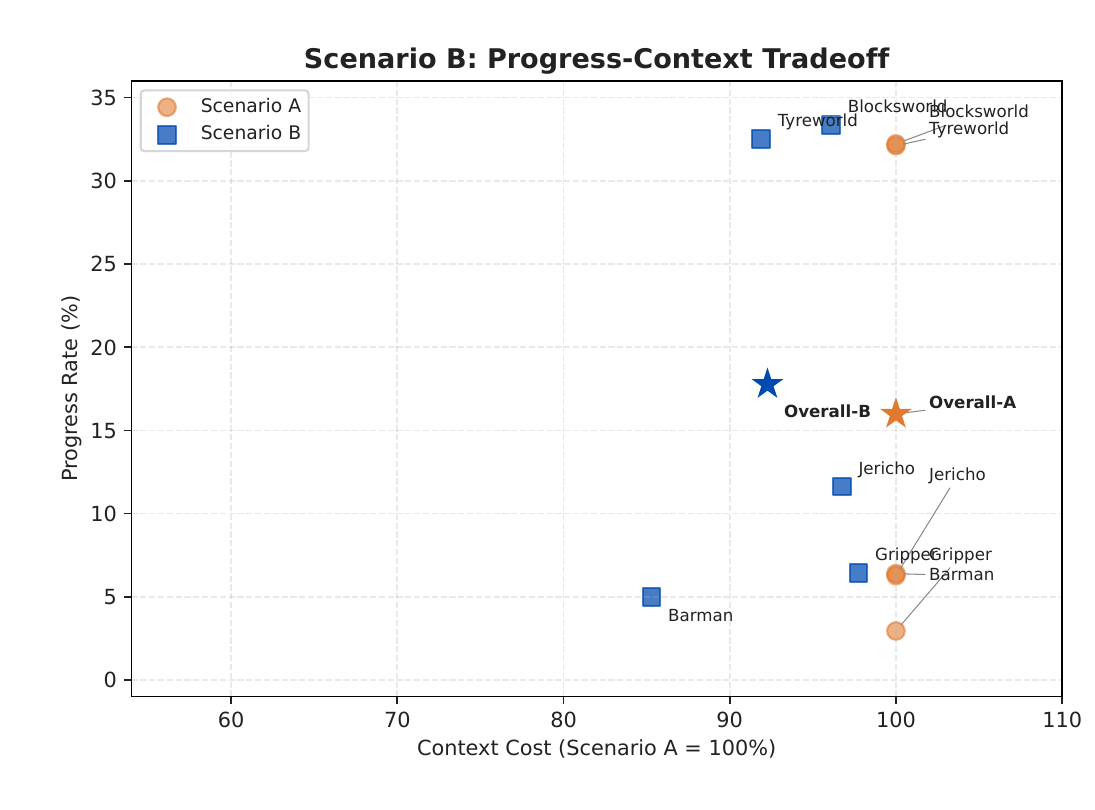}
    \caption{Scenario B.}
    \label{fig:scenario_b_pr_context_subfig}
  \end{subfigure}
  \caption{Progress rate vs.\ context token consumption.}
  \label{fig:scenario_a_pr_context}
\end{figure}

Figure~\ref{fig:scenario_a_pr_context} plots the progress rate against context token consumption for each task, where asterisks denote averages over all environments. In Scenario A (Figure~\ref{fig:scenario_a_pr_context_subfig}), \textsc{CoMIC} improves task progress while reducing context cost relative to the \textsc{Standard} baseline. In Scenario B (Figure~\ref{fig:scenario_b_pr_context_subfig}), guided by experience distilled from the stronger edge agent's trajectories, the weak agent further improves progress on most tasks while reducing context consumption relative to Scenario A.

\subsection{Additional Cloud-side Analysis}
\label{app:cloud-metrics}
This section provides detailed coordination statistics, experience pipeline metrics, and total system overhead analysis for the cloud layer.

\begin{table}[!htbp]
\caption{Cloud-side coordination statistics in Scenarios A and B.}
\label{tab:cloud-coordination}
\centering
\small

\definecolor{AvgBlue}{RGB}{110,165,218}
\definecolor{AvgPurple}{RGB}{179,138,208}
\colorlet{AvgBlueLight}{AvgBlue!25!white}
\colorlet{AvgPurpleLight}{AvgPurple!25!white}

\newcommand{\avgrowblue}{%
  \tikz[remember picture,overlay]
    \fill[AvgBlueLight]
      (-0.15em,-0.36em) rectangle (\linewidth,0.9em);%
}

\newcommand{\avgrowpurple}{%
  \tikz[remember picture,overlay]
    \fill[AvgPurpleLight]
      (-0.15em,-0.36em) rectangle (\linewidth,0.9em);%
}

\renewcommand{\arraystretch}{0.92}
\setlength{\tabcolsep}{0pt}

% Coordination Statistics for Scenario A and Scenario B
\begin{tabular*}{\linewidth}{@{\extracolsep{\fill}}lrrrrr@{}}
\toprule
\textbf{} &
\textbf{Hit $\uparrow$} &
\textbf{Conf. $\uparrow$} &
\textbf{Lat. $\downarrow$} &
\textbf{Ins. $\uparrow$} &
\textbf{ACK $\uparrow$} \\
\midrule

\textbf{\textit{PDDL}} & & & & & \\
\textsc{Scenario A} & 44.12 & 0.609 & 73.71 & 8.42 & 76.25 \\
\textsc{Scenario B} & 26.28 & 0.473 & 84.35 & 4.71 & 50.00 \\
\avgrowblue\textsc{Avg.} & 35.20 & 0.541 & 79.03 & 6.57 & 63.13 \\
\midrule

\textbf{\textit{Jericho}} & & & & & \\
\textsc{Scenario A} & 49.15 & 0.553 & 48.56 & 1.25 & 50.00 \\
\textsc{Scenario B} & 12.24 & 0.399 & 7.66  & 1.00 & 0.00 \\
\avgrowblue\textsc{Avg.} & 30.70 & 0.476 & 28.11 & 1.13 & 25.00 \\
\midrule

\textbf{\textit{Overall}} & & & & & \\
\textsc{Scenario A} & 45.12 & 0.597 & 68.68 & 6.98 & 71.00 \\
\textsc{Scenario B} & 23.47 & 0.459 & 69.01 & 3.97 & 40.00 \\
\avgrowpurple\textsc{Avg.} & 34.30 & 0.528 & 68.85 & 5.48 & 55.50 \\
\bottomrule
\end{tabular*}

\renewcommand{\arraystretch}{1.0}
\end{table}

According to the detailed numerical comparisons in Table~\ref{tab:cloud-coordination}, Scenario~A outperforms Scenario~B across key metrics. This is not because the cloud generates guidance of intrinsically higher value in Scenario~A. Rather, because the edge agents in this scenario use the same base model, their generated trajectories are similar at the abstract level. Consequently, the global guidance selected by the cloud critic is more applicable and easier for these base models to understand and execute.

In the mixed-strength setting (Scenario~B), the guidance selected by the cloud critic incorporates more abstract patterns from the trajectories of strong agents. Since the weak edge agent struggles to plan subtasks over a long horizon as its stronger counterparts do, the subgoals it plans often fail to match those under which global guidance is indexed, preventing it from retrieving the corresponding guidance. This leads to the decline in the \textbf{Hit} and \textbf{ACK} rates. However, as discussed in \S\ref{sec:hetero-analysis}, when the weak agent in Scenario~B does plan the correct subgoals, its episode completion metrics improve. This explains why the weak LLM agent in Scenario~B ultimately achieves better edge-side performance than that in Scenario~A, despite the drop in cloud-side coordination efficiency.

\subsection{Additional Cloud-Edge Collaboration Analysis}
\label{app:collab-metrics}
\begin{figure}[!htbp]
  \centering
  \includegraphics[width=\linewidth]{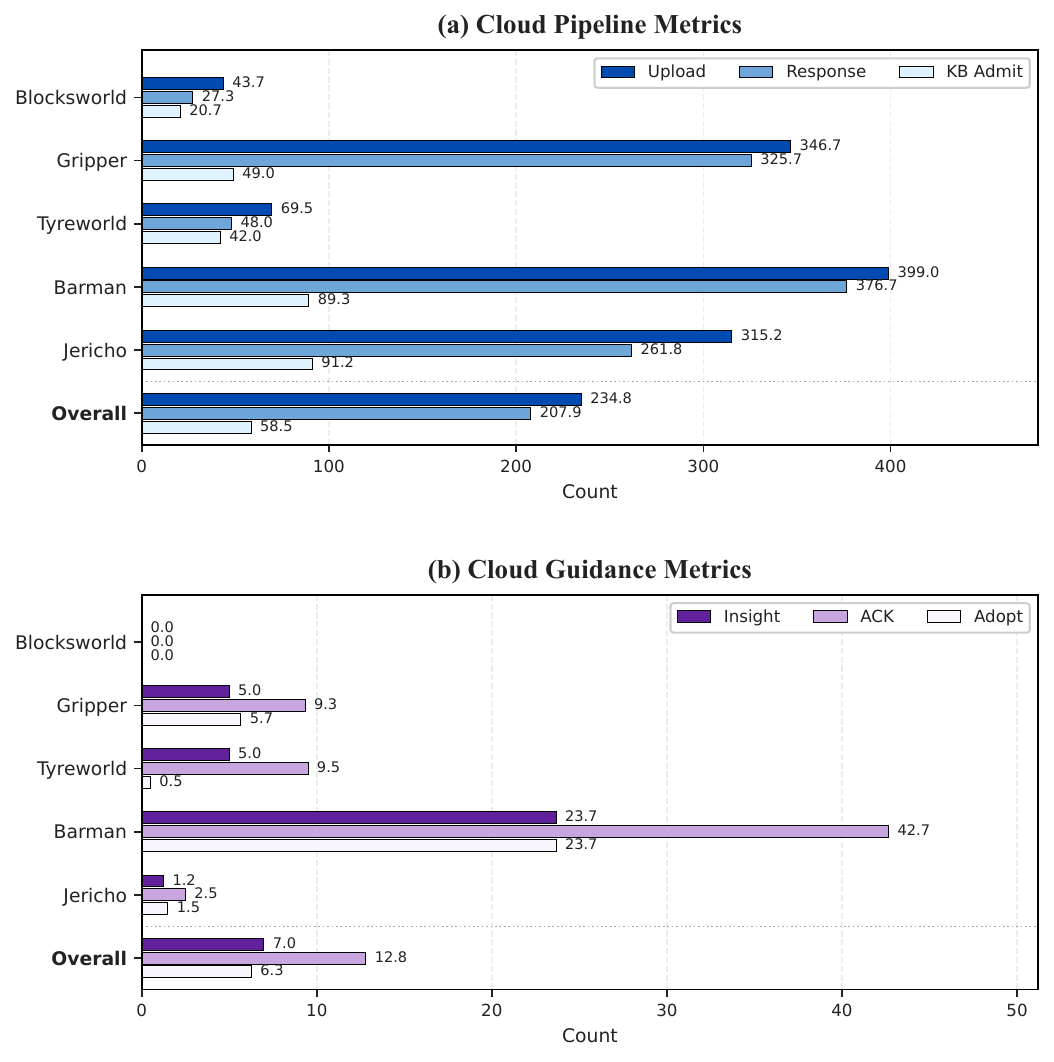}
  \caption{Cloud critic metrics in Scenario A.}
  \label{fig:scenario_a_cloud_metrics}
\end{figure}

\begin{figure}[!htbp]
  \centering
  \includegraphics[width=\linewidth]{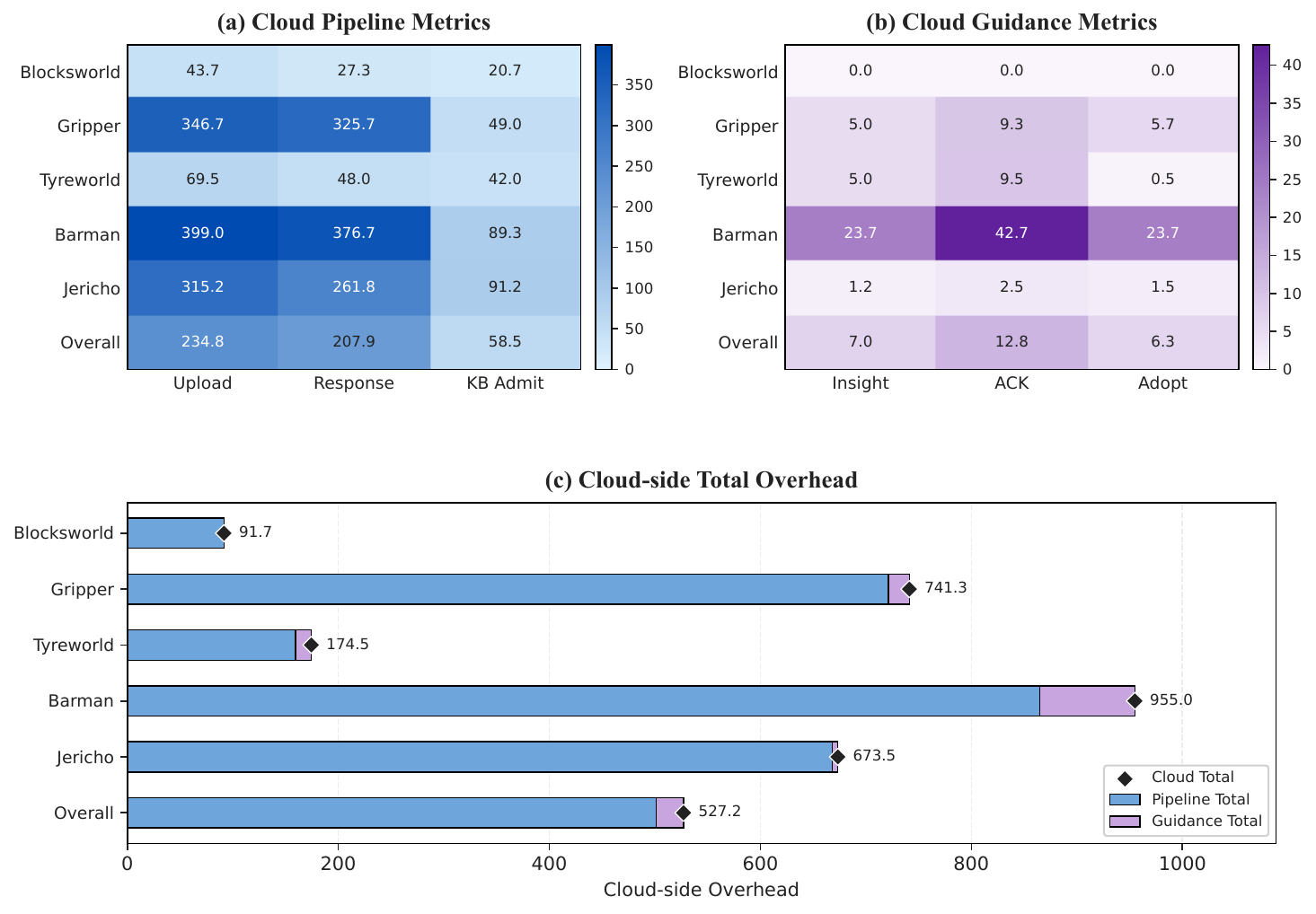}
  \caption{Cloud overhead metrics in Scenario A.}
  \label{fig:scenario_a_cloud_overhead}
\end{figure}

\begin{figure}[!htbp]
  \centering
  \includegraphics[width=\linewidth]{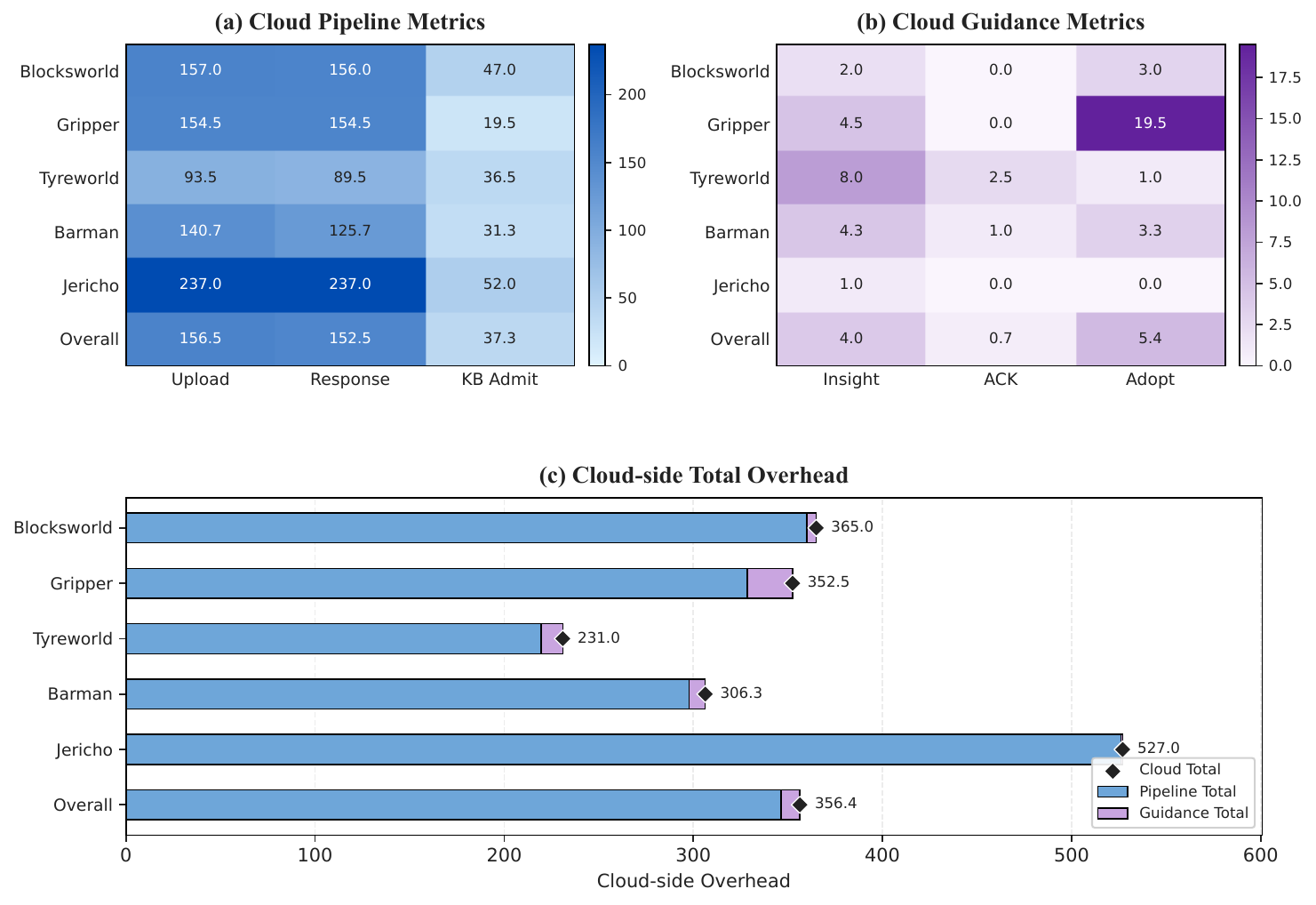}
  \caption{Cloud overhead metrics in Scenario B.}
  \label{fig:scenario_b_cloud_overhead}
\end{figure}

Figures~\ref{fig:scenario_a_cloud_metrics} and~\ref{fig:scenario_b_cloud_metrics} present the cloud critic metrics defined in Appendix~\ref{app:metrics} for each task in Scenarios~A and~B, respectively. They track the lifecycle from experience upload to guidance adoption through the cloud pipeline metrics (Upload, Response, and KB Admit) and the cloud guidance metrics (Insight, ACK, and Adopt). Figures~\ref{fig:scenario_a_cloud_overhead} and~\ref{fig:scenario_b_cloud_overhead} report the corresponding resource consumption of the cloud-side memory mechanism: panels (a) and (b) show the distillation pipeline cost (Pipeline Total) and the guidance retrieval cost (Guidance Total), and panel (c) shows their sum (Cloud Total) together with the proportion of each component.

\begin{figure}[!htbp]
  \centering
  \includegraphics[width=\linewidth]{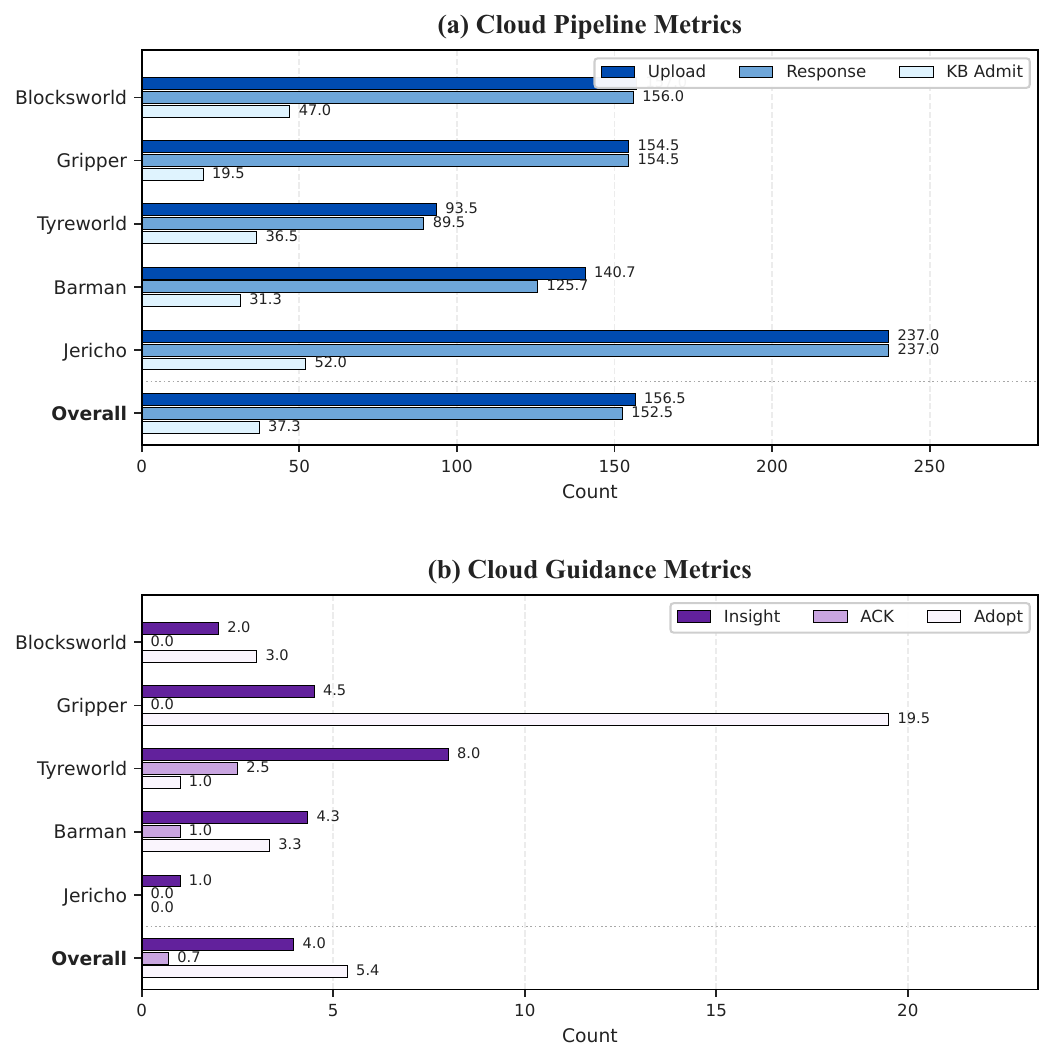}
  \caption{Cloud critic metrics in Scenario B.}
  \label{fig:scenario_b_cloud_metrics}
\end{figure}

\clearpage

\end{document}